\documentclass[11pt]{article}
\usepackage{psfig}

\newenvironment{eq}{\begin{equation}}{\end{equation}}
\newcommand{\beq}{\begin{eq}}
\newcommand{\eeq}{\end{eq}}

\newenvironment{eqa}{\begin{eqnarray}}{\end{eqnarray}}
\newcommand{\beqa}{\begin{eqa}}
\newcommand{\eeqa}{\end{eqa}}

\newenvironment{eqs}{\begin{eqnarray*}}{\end{eqnarray*}}
\newcommand{\beqs}{\begin{eqs}}
\newcommand{\eeqs}{\end{eqs}}

\newenvironment{dm}{\begin{displaymath}}{\end{displaymath}}
\newcommand{\bdm}{\begin{dm}}
\newcommand{\edm}{\end{dm}}

\newenvironment{mat}[1]{\left[\begin{array}{#1}}{\end{array}\right]}
\newcommand{\bmx}[1]{\begin{mat}{#1}}
\newcommand{\emx}{\end{mat}}

\newenvironment{dt}[1]{\left|\begin{array}{#1}}{\end{array}\right|}
\newcommand{\bdt}[1]{\begin{dt}{#1}}
\newcommand{\edt}{\end{dt}}

\newcommand{\gssa}[4]{\mbox{\boldmath $#1$}_{#2}^{#3}(#4)}
\newcommand{\ga}[2]{\mbox{\boldmath $#1$}(#2)}
\newcommand{\gss}[3]{\mbox{\boldmath $#1$}_{#2}^{#3}}

\newcommand{\hgss}[3]{\hat{\mbox{\boldmath $#1$}}_{#2}^{#3}}
\newcommand{\hga}[2]{\hat{\mbox{\boldmath $#1$}}(#2)}

\newcommand{\sr}[2]{\stackrel{#1}{#2}}

\newcommand{\pref}[1]{(\ref{#1})}

\newcounter{alpha}

\title{Blind Source Separation: Fundamentals and Recent Advances\\
{\large (A tutorial overview presented at SBrT-2001)}}
\author{\it Eleftherios Kofidis}
\date{Department of Statistics and Insurance Science\\University
of Piraeus\\185~34 Piraeus, Greece.\\E-mail: {\tt
kofidis@unipi.gr}} 
\begin{document}
\maketitle
\begin{abstract}
Blind source separation (BSS), i.e., the decoupling of unknown
signals that have been mixed in an unknown way, has been a topic
of great interest in the signal processing community for the last
decade, covering a wide range of applications in such diverse
fields as digital communications, pattern recognition, biomedical
engineering, and financial data analysis, among others. This
course aims at an introduction to the BSS problem via an
exposition of well-known and established as well as some more
recent approaches to its solution. A unified way is followed in
presenting the various results so as to more easily bring out
their similarities/differences and emphasize their relative
advantages/disadvantages. Only a representative sample of the
existing knowledge on BSS will be included in this course. The
interested readers are encouraged to consult the list of
bibliographical references for more details on this exciting and
always active research topic.
\end{abstract}

\newpage
\section{Introduction}
Consider the following scenario. A number of people are found in
a room and involved in loud conversations in groups, just as it
would happen in a cocktail party. There might also be some
background noise, which could be music, car noise from outside,
etc. Each person in this room is therefore forced to listen to a
mixture of speech sounds coming from various directions, along
with some noise. These sounds may come directly to one's ear or
have first suffered a sequence of reverberations because of their
reflections on the room's walls. The problem of focusing one's
listening attention on a particular speaker among this cacophony
of conversations and noise has been known as the {\em cocktail
party problem} \cite{a92}. It consists of separating a mixture of
speech signals of different characteristics with noise added to
it. The signals are a-priori unknown (one listens only to a
combination of them) as is also the way they have been mixed. The
above scenario is a good analog for many other examples of
situations that demand for a separation of mixed signals with no
presupposed knowledge on the signals and the system mixing them.
The list is long:
\begin{itemize}
\item Reception for single- and multi-user communications with no
training data \cite{d97,tlh98,m98,j00}
\item Analysis of biomedical signals (e.g.,
electroencephalographic (EEG) \cite{l98} and electrocardiographic
(ECG) signals \cite{bmo98}, fetal electrocardiography
\cite{lmv00f}, functional Magnetic Resonance Imaging (fMRI) data
\cite{dhg00}, etc.)
\item Restoration of images taken from inaccessible scenes \cite{kh96} (e.g.,
ultrasonography \cite{apr95}, astronomical imaging \cite{b82},
images of a nuclear reactor, etc.)
\item Feature extraction \cite{bs96,hohh98}
\item Uncalibrated (or partially calibrated) array processing
\cite{car90a,car91b}
\item Geophysical exploration \cite{lac97}
\item Denoising \cite{hnb01}
\item Voice-controlled machines \cite{tlsh91}
\item Circuit testing \cite{tlsh91}
\item Semiconductor manufacturing \cite{tlsh91}
\item $\cdots$
\end{itemize}

This kind of problem is commonly referred to as {\em Blind Source
Separation (BSS).} The term {\em blind} is used to emphasize that
the signals are to be separated on the basis of their mixture
only, without accessing the signals themselves and/or knowing the
mixing system, i.e., blindly. Additional information is usually
required, involving however only general statistical and/or
structural properties of the sources and the mixing mechanism.
For example, in a multi-user digital communications system one
might assume the signals transmitted to be mutually independent,
i.i.d., and BPSK modulated, and the channel(s) to be linear,
time-invariant, but otherwise unknown. If no training data are
allowed to be transmitted along, due for example to bandwidth
limitations, then reception has to be performed blindly.

BSS sounds and is indeed a difficult problem, especially in its
most general version. Note that humans experience in general not
much difficulty in coping successfully with the cocktail party
problem. However, this fact should not be overemphasized, in view
of the amazing complexity and performance of the human cognitive
system. It would not be that easy to separate speakers talking at
the same time by relying, for example, solely on the recordings
of a few microphones placed in the room. Nevertheless, a lot of
progress has been made in the last ten years or so in the study
of the BSS problem and a great number of successful methods, both
general and specialized, have been developed and their
applicability demonstrated in several contexts. Self-organizing
neural networks \cite{g99} have, as it will be seen in this
course, directly or indirectly played an important role in this
research effort, exploiting in a way the human capabilities for
BSS through their analogy with related biological mechanisms.

This course aims at providing a tutorial overview of the most
important aspects of the BSS problem and approaches to solving
it. For reasons to be explained, only methods based on
higher-order statistics are discussed. In addition to well-known
and established results and algorithms, some more recent
developments are also presented that show promise in
improving/extending the capabilities and removing the limitations
of earlier, classical approaches. BSS is a vast topic, impossible
to fit into a short course. Only a selected, yet representative,
part of it is thus treated here, hopefully sufficient to motivate
interested readers to further explore it with the aid of the
(necessarily incomplete) provided bibliography.

The rest of this text is organized as follows. The BSS problem is
stated in precise mathematical terms in Section~2, along with a
set of assumptions that allow its solution. Section~3 examines
the possibility of solving the problem by relying on second-order
statistics (SOS) only. It is argued there that SOS are
insufficient to provide a piercing solution in all possible
scenarios. Higher-order statistics (HOS) are generally required.
An elegant and intuitively pleasing manner of expressing and
exploiting this additional information is provided by
information-theoretic (IT) criteria. Information is defined in
Section~4 and its relevance to the BSS problem is explained. The
criterion of minimal mutual information (MI) is formulated and
interpreted in terms of the Kullback-Leibler divergence. A
unification of seemingly unrelated IT and statistical criteria is
put forward in Section~5, based on the criterion of minimum MI.
This results in a single, ``universal" criterion, that is shown
to be interpretable in terms of the so-called nonlinear Principal
Component Analysis (PCA). A corresponding algorithmic scheme,
resulting from its gradient optimization, is presented in
Section~6, along with its interpretation as a nonlinear-Hebbian
learning scheme and a discussion of the choice of the involved
nonlinearities and their implications to algorithm's stability.
It is also shown how the use of the relative gradient in the above
algorithm transforms it to one that enjoys the so-called
equivariance property. In Section~7 the close relation of the
above scheme with the recently proposed FastICA algorithm is
demonstrated. The latter is shown in turn to specialize to the
super-exponential algorithm, and its variant, the
constant-modulus algorithm, well-known from the channel
equalization problem. Section~8 discusses algebraic methods for
BSS, based on the concept of cumulant tensor diagonalization.
Their connection with the above approaches is pointed out, via
their interpretation as cost-optimization schemes. Some recently
reported deterministic methods, that have been demonstrated to
work for short data lengths are also included. The model employed
thus far involves a static (memoryless), linear, time-invariant
and perfectly invertible mixing system with stationary sources.
These simplifying assumptions are relaxed in the following
sections. Non-invertible mixing systems are dealt with in
Section~9, where both algebraic and criterion-based approaches
are discussed. Dynamic (convolutive) mixtures are discussed in
Section~10, where methods for dealing with nonlinear or
time-varying mixtures and nonstationary sources are also briefly
reviewed. Some concluding remarks are included in Section~11.

\noindent {\bf Notation:} Bold lower case and upper case letters
will designate vectors and matrices, respectively. Unless stated
otherwise,  capital, calligraphic, boldface letters will denote
higher-order tensors. $\gss{I}{m}{}$ will be the $m\times m$
identity matrix. The subscript $m$ will be omitted whenever
understood from the context. The superscript $^{T}$ will denote
transposition. $\otimes$ is the Kronecker product. The inverse
operation of building a vector from a matrix by stacking its
columns one above another will be denoted by ${\rm unvec}(\cdot).$

\section{Problem Statement}
Consider the mapping \beq \ga{u}{n}={\cal
F}(\ga{a}{n},\ga{v}{n},n) \label{eq:u=f(a)} \eeq mixing the $N$
{\em source signals} \beq \ga{a}{n}=\bmx{cccc} a_{1}(n) &
a_{2}(n) & \cdots & a_{N}(n)\emx^{T} \eeq
 and the $K$ {\em noise signals}
\beq \ga{v}{n}=\bmx{cccc} v_{1}(n) & v_{2}(n) & \cdots & v_{K}(n)
\emx^{T} \eeq
 via the {\em mixing system} ${\cal F}(\cdot,\cdot,\cdot)$
to yield the {\em mixture} \beq \ga{u}{n}=\bmx{cccc} u_{1}(n) &
u_{2}(n) & \cdots & u_{M}(n) \emx^{T}. \eeq $\cal F$ is in general
{\em nonlinear} and its time argument, $n$, signifies that it can
also be time-varying. BSS can thus be stated as the problem of
identifying the mapping $\cal F$ and/or extracting the sources
$\gss{a}{}{}$ having access only to the mixture $\gss{u}{}{}.$
Both the system $\cal F$ and its inputs $\gss{a}{}{},\gss{v}{}{}$
are assumed unknown. Nevertheless, so that the problem have a
solution, some general structural and/or statistical properties
of the quantities involved are assumed. The mixing system is
usually supposed to be linear and time-invariant. The most common
assumption for the sources is that they are mutually independent
and independent of the noise. Depending on the application
context, more information may be available. For example, in a
digital communications system the source signals might be known
to belong to some discrete alphabet. The set of assumptions made
for $\cal F$ and $\gss{a}{}{},\gss{v}{}{}$ constitute the {\em
model} adopted for the problem and determine the range of possible
solution approaches.

Due to its relative simplicity and wide applicability, the case
of a linear, time-invariant mixing system (with stationary
sources) has been the focus of the great majority of the works on
BSS. This can be written as \beq
\ga{u}{n}=\gss{H}{}{}\ga{a}{n}+\ga{v}{n} \label{eq:u=Ha+v} \eeq
where $\gss{H}{}{}$ is an $M\times N$ {\em mixing matrix} and
$K=M.$ The following set of assumptions is typical:
\begin{itemize}
\item[A1.] The signals $a_{i}(n)$ and $v_{i}(n)$ are stationary and
zero-mean.
\item[A2.] The sources $a_{i}(n)$ are statistically independent.
\item[A3.] The noises $v_{i}(n)$ are statistically independent and
independent of the sources.
\item[A4.] The number of {\em sensors} exceeds or equals the number of
sources: $M\geq N.$
\end{itemize}
The mixing system in \pref{eq:u=Ha+v} is moreover static, i.e.,
memoryless. This model could also encompass dynamic systems if it
included an assumption of independence between samples of a
signal $a_{i}(n)$ and a structural constraint (Toeplitz) on
$\gss{H}{}{}.$ Linear systems with memory will be briefly
addressed in Section~10.

The goal of the {\em linear} BSS is to determine a {\em
separating matrix} $\gss{G}{}{}$ such that \beq
\ga{y}{n}=\gss{G}{}{}\ga{u}{n} \label{eq:y=Gu} \eeq is a good
approximation of the source signals. Assumption~A4 signifies that
the mixture is not under-determined, i.e., in the absence of
noise, if $\gss{H}{}{}$ is identified and of full rank, then it
can be inverted to yield the sources. If noise is not weak enough
to be neglected, perfect recovery of all sources is impossible. In
fact, the noisy mixture may be viewed as a special case of a
noise-free mixture that does not meet A4. Just write
\pref{eq:u=Ha+v} in the form
\[
\ga{u}{n}=\gss{H}{}{\prime}\gssa{a}{}{\prime}{n}
\]
where $\gss{H}{}{\prime}=\bmx{cc} \gss{H}{}{} & \gss{I}{M}{}
\emx$ and $\gssa{a}{}{\prime}{n}=\bmx{cc} \gssa{a}{}{T}{n} &
\gssa{v}{}{T}{n} \emx^{T}.$ Then $\gss{H}{}{\prime}$ is $M\times
(M+N)$ and cannot be (linearly) inverted.\footnote{Possible ways
for recovering the sources in such under-determined mixtures, via
nonlinear inversion or one-by-one source extraction, will be
discussed in Section~9.} Hence, one can, without loss of
generality, write \beq \ga{u}{n}=\gss{H}{}{}\ga{a}{n}
\label{eq:u=Ha} \eeq for the linear mixing process. The general
linear BSS setup is shown schematically in Fig.~1.
\begin{figure}
\centerline{\psfig{figure=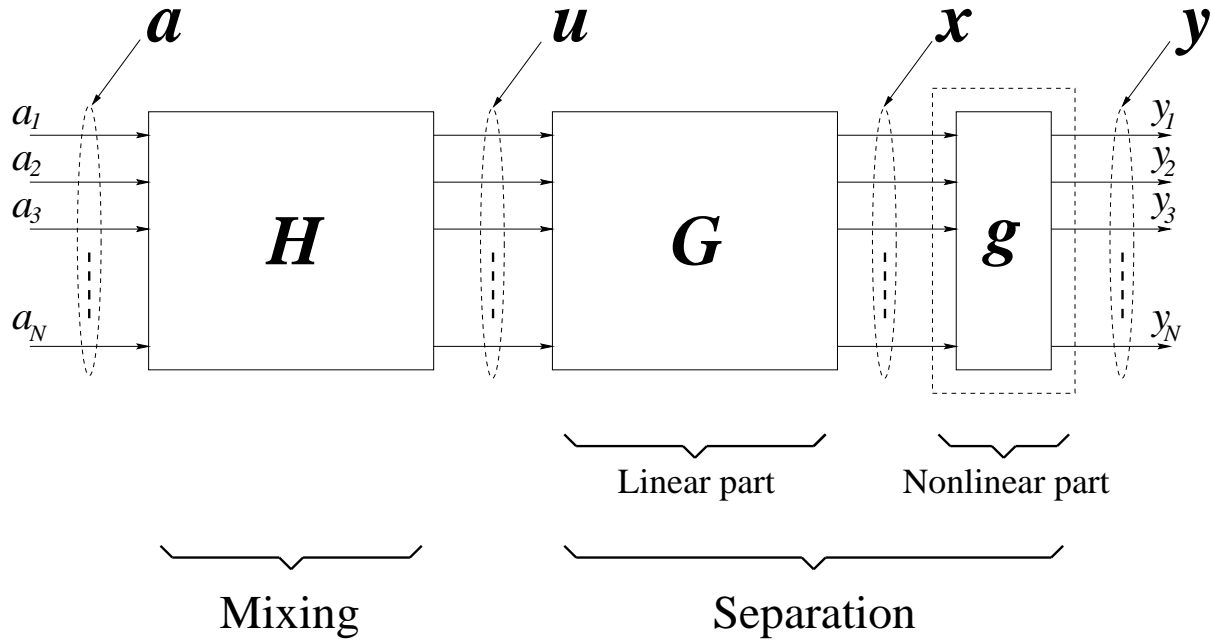,width=16 true cm}}
\caption{Linear BSS setup.}
\end{figure}
The (nonlinear) function $\ga{g}{\cdot}$ at the separator's
output may not be really there. It is needed however in the
development of the criteria and associated algorithms.

It must be noted that, unless additional information is
available, there is an indeterminacy with respect to the scaling
and order of the separated sources \cite{tlsh91}. The result of
the mixing in \pref{eq:u=Ha} will not change if a source is
multiplied with a scalar and the corresponding column of
$\gss{H}{}{}$ is divided by it. A permutation of the source
signals accompanied by an analogous permutation of the columns of
the mixing matrix will not have any effect on $\ga{u}{n}$ either.
The best one could expect therefore is to recover $a_{i}$'s
scaled and/or permuted, i.e.,
\[
\ga{y}{n}=\gss{P}{}{}\gss{D}{}{}\ga{a}{n}
\]
where $\gss{P}{}{}$ is a permutation matrix and $\gss{D}{}{}$ is
a diagonal (invertible) matrix. The following assumption on the
source energy can therefore be made without harming the
generality:
\begin{itemize}
\item[A5.] The sources have unit variance: $E[|a_{i}(n)|^{2}]=1.$
\end{itemize}

This course, in its main part, will focus on the noise-free,
linear, static (instantaneous) mixture case \pref{eq:u=Ha}, with
as many sensors as sources, i.e., $M=N$, an invertible mixing
matrix, and sources satisfying assumptions A1, A2, A5.
Furthermore, to simplify the presentation, we shall frequently
omit the inherent scale and order indeterminacy and write
$\gss{G}{}{}=\gss{H}{}{-1}.$ Moreover, all quantities are assumed
to be real. It will be usually straightforward to extend the
results to the complex case.

The model adopted may not be realistic in many practical
scenarios. It is sufficient however to convey the basic ideas
underlying the problem, while keeping things simple. Besides, as
the following analysis should show, its study is already far from
being trivial.

\section{Are Second-Order-Statistics Sufficient?}

Based on the model assumptions made above, the BSS problem is
solved once the mixing matrix $\gss{H}{}{}$ has been identified.
Thus the question arises whether $\gss{H}{}{}$ could be somehow
extracted from measuring appropriate quantities related to the
mixture signals. Note that the autocorrelation matrix of the
observation vector $\ga{u}{n}$ is given by: \beq
\gss{R}{u}{}=E[\ga{u}{n}\gssa{u}{}{T}{n}]=\gss{H}{}{}\gss{R}{a}{}\gss{H}{}{T}=
\gss{H}{}{}\gss{H}{}{T} \label{eq:Ru=HH^T} \eeq where the index
$n$ is omitted due to stationarity and the source autocorrelation
matrix $\gss{R}{a}{}=E[\ga{a}{n}\gssa{a}{}{T}{n}]$ equals the
identity according to the assumptions A1, A2, and A5. The above
equation suggests that the mixing matrix might be identifiable as
a square root of $\gss{R}{u}{},$ i.e.,
$\gss{H}{}{}=\gss{R}{u}{1/2}$, via its eigenvalue decomposition
(EVD) for example. In fact it is only a basis for the column space
of $\gss{H}{}{}$ that this would provide.\footnote{If
$\gss{H}{}{}$ results as a convolution matrix of a filter, this
information may suffice to identify it \cite{lmr01}.} It is easy
to see that not only $\gss{H}{}{}$ but also any product of it
with an orthogonal matrix, $\gss{H}{}{}\gss{Q}{}{T}$, will verify
\pref{eq:Ru=HH^T}. In other words, $\gss{R}{u}{}$ allows
$\gss{H}{}{}$ to be identified only up to an orthogonal factor,
$\gss{Q}{}{T}$.

It is instructive to observe that this ambiguity is an instance
of the well-known phase indeterminacy problem in spectral
factorization for linear prediction \cite{p84}, where the role of
the sources play the innovation sequence samples. The net effect
of the corresponding ``separator" $\gss{T}{}{}=\gss{R}{u}{-1/2}$
is to {\em whiten} the observations: \beq
\ga{\bar{u}}{n}=\gss{T}{}{}\ga{u}{n} \label{eq:PCA} \eeq with
\[
\gss{R}{\bar{u}}{}=\gss{T}{}{}\gss{R}{u}{}\gss{T}{}{T}=\gss{I}{N}{}.
\]
The operation \pref{eq:PCA} amounts to a projection of the
available data $\ga{u}{n}$ onto the principal directions
determined from $\gss{R}{u}{}$ and is known in the statistics
jargon as {\em Principal Component Analysis (PCA)}
\cite{cb89,ep98}. The whitened vector $\ga{\bar{u}}{n}$, whose
entries are the principal components of $\ga{u}{n}$, is commonly
referred to as {\em standardized} or {\em sphered}
\cite{lac97,cch00}.

The above reduces the original problem to one where the mixing
matrix is constrained to be orthogonal: \beq
\ga{\bar{u}}{n}=\gss{T}{}{}\gss{H}{}{}\ga{a}{n}=\gss{Q}{}{}\ga{a}{n},
\mbox{\ \ } \gss{Q}{}{}\gss{Q}{}{T}=\gss{I}{}{}. \label{eq:u=Qa}
\eeq It is clear that to eliminate the remaining uncertainty
about $\gss{H}{}{}$ more information is required. In fact it
turns out that, subject to some hypotheses on the source signals
to be described shortly, the mixing matrix can be identified by
exploiting the mixture autocorrelation for a nonzero lag as well.
Assume that there is a lag $k\neq 0$ for which no two sources
have the same autocorrelation, i.e., \beq
E[a_{i}(n)a_{i}(n-k)]\neq E[a_{j}(n)a_{j}(n-k)], \mbox{\ for all\
} i\neq j. \label{eq:aiaineqajaj} \eeq Take the lag-$k$
autocorrelation matrix of the standardized data: \beq
\gssa{R}{\bar{u}}{}{k}=E[\ga{\bar{u}}{n}\gssa{\bar{u}}{}{T}{n-k}]=
\gss{Q}{}{}\gssa{R}{a}{}{k}\gss{Q}{}{T}. \label{eq:Ru(k)} \eeq By
A2 $\gssa{R}{a}{}{k}$ is diagonal, hence the above equation
represents an EVD of $\gssa{R}{\bar{u}}{}{k}.$ In view of
\pref{eq:aiaineqajaj}, the eigenvalues $E[a_{i}(n)a_{i}(n-k)]$
are distinct. This implies that the columns of $\gss{Q}{}{}$ are
uniquely determined from an EVD of $\gssa{R}{\bar{u}}{}{k}$
subject to sign/order changes.

The above method, known as {\em Algorithm for Multiple Unknown
Signals Extraction (AMUSE)} \cite{tlsh91}, is found at the heart
of many other BSS algorithms relying on the mixture's SOS. These
approaches provide simple and fast BSS tools whose effectiveness
is well documented in the relevant literature \cite{tlsh91}. They
rely heavily however on the assumption that not two sources have
the same power spectral contents (see \pref{eq:aiaineqajaj}). If
two sources have different, but similar, power spectra, the
separation will be theoretically possible but with practically
unreliable results \cite{lac97}. Moreover, it is clear that these
methods will not apply to scenarios where $a_{i}(n)$ should be
considered as white, and even i.i.d.\footnote{The latter source
model is very common in digital communications \cite{hay94}.}

Since it is our purpose to cover methods for the BSS problem in
its generality as described in Section~2, no assumption like
\pref{eq:aiaineqajaj} will be made henceforth. It is then made
clear from the above discussion that \pref{eq:u=Qa} represents
the limit of what a SOS-based approach can bring to our problem.
In fact, one cannot go any further than that in the case that
sources are Gaussian, since such a signal is completely described
by its mean and covariance. It will be shown later on that at
most one source can be Gaussian if the BSS problem is to have any
solution at all. Observe that in the above algorithm the sources
did not have to be independent, only the fact that they are
uncorrelated was made use of. This suggests that confining
ourselves to the second-order information does not exploit all
available knowledge. It is of interest to recall that it is for a
Gaussian signal that the notions of independence and
uncorrelatedness coincide \cite{p84}. However, most of the sources
encountered in practical applications, e.g., speech, music, data,
image, are distinctly non-Gaussian. It follows therefore that
another transformation, that can extract from the data more
structure than PCA is able to, is needed. This will be what we
call {\em Independent Component Analysis (ICA).}

\section{Mutual Information and Independent Component Analysis}
\subsection{Statistical Independence}
The independence assumption for the source signals can be
equivalently stated as the equality of the joint probability
density function (pdf) of the source vector $\ga{a}{n}$ with the
product of the marginal pdf's of its entries: \beq
p_{a}(\gss{a}{}{})=\tilde{p}_{a}(\gss{a}{}{})\sr{\triangle}{=}
\prod_{i=1}^{N}p_{a_{i}}(a_{i}). \label{eq:indepa} \eeq Hence an
analogous equality would hold for the output of a successful
separating system, i.e., \beq
p_{y}(\gss{y}{}{})=\tilde{p}_{y}(\gss{y}{}{})\sr{\triangle}{=}\prod_{i=1}^{N}p_{y_{i}}(y_{i}).
\label{eq:indepy} \eeq The above represents the goal of the ICA
transformation, i.e., (linearly) transforming a random vector
into one with (as) independent components (as possible).
Information theory constitutes a powerful and elegant framework
for expressing and studying this problem.

\subsection{Entropy and Mutual Information}
Recall that the {\em entropy} of a random variable provides a
measure of our uncertainty about the value this variable takes.
For a continuous-valued random vector $\gss{x}{}{}$, it takes the
form\footnote{We will be concerned with the Shannon entropy. For
IT approaches to BSS that employ other definitions of entropy we
refer the reader to \cite{pxf00}.} \cite{p84,hay98} \beq
H(\gss{x}{}{})=-E[\ln p_{x}(\gss{x}{}{})]=-\int
p_{x}(\gss{x}{}{})\ln p_{x}(\gss{x}{}{})d\gss{x}{}{}.
\label{eq:H(x)}\eeq In fact this is rather what we call the {\em
differential entropy} of $\gss{x}{}{}$. The entropy of a
continuous random vector goes to infinity (since the uncertainty
on the value assumed is infinite) and is equal to the
differential entropy only modulo a reference term \cite{hay98}.
However, since it will be the optimization of such quantities and
mainly their differences that will be interested in, we may rest
on the differential entropy and call it simply entropy from now
on.

$H(\gss{x}{}{})$ is a measure of the uncertainty on the value of
the vector $\gss{x}{}{}.$ Let another random vector
$\gss{y}{}{}$, and let $p_{x|y}(\gss{x}{}{}|\gss{y}{}{})$ denote
the pdf of $\gss{x}{}{}$ conditioned on $\gss{y}{}{}.$ Then a
measure of the uncertainty remaining about $\gss{x}{}{}$ after
having observed $\gss{y}{}{}$ can be given by the {\em
conditional entropy}: \beq H(\gss{x}{}{}|\gss{y}{}{})=-E[\ln
p_{x|y}(\gss{x}{}{}|\gss{y}{}{})]=-\int
p_{x,y}(\gss{x}{}{},\gss{y}{}{})\ln
p_{x|y}(\gss{x}{}{}|\gss{y}{}{})d\gss{x}{}{}d\gss{y}{}{}
\label{eq:H(x|y)} \eeq where \beq
p_{x,y}(\gss{x}{}{},\gss{y}{}{})=p_{x|y}(\gss{x}{}{}|\gss{y}{}{})p_{y}(\gss{y}{}{})
\label{eq:p(x,y)} \eeq is the joint pdf of $\gss{x}{}{}$ and
$\gss{y}{}{}.$ It is then reasonable to say that the difference
\beq
I(\gss{x}{}{},\gss{y}{}{})=H(\gss{x}{}{})-H(\gss{x}{}{}|\gss{y}{}{})
\label{eq:I(x,y)} \eeq represents the {\em information} concerning
$\gss{x}{}{}$ that is acquired when observing $\gss{y}{}{}.$ It
is termed as the {\em mutual information (MI)} between
$\gss{x}{}{}$ and $\gss{y}{}{}.$ As it will be seen shortly,
$I(\gss{x}{}{},\gss{y}{}{})$ is always nonnegative and vanishes
if and only if $\gss{x}{}{}$ and $\gss{y}{}{}$ are statistically
independent. This is to be expected since for independent
$\gss{x}{}{}$ and $\gss{y}{}{}$ the observation of the one does
not provide any information concerning the other. This is also
seen in the definition \pref{eq:I(x,y)} since then
$H(\gss{x}{}{}|\gss{y}{}{})=H(\gss{x}{}{}).$ The MI is thus a
meaningful measure of statistical dependence and, in fact, it
will prove to be a good starting point for a fruitful study of
ICA.

\subsection{Kullback-Leibler Divergence}
The condition for independence stated in \pref{eq:indepy} implies
that independence can also be viewed in terms of the distance
between $p_{y}(\gss{y}{}{})$ and $\tilde{p}_{y}(\gss{y}{}{}).$ A
measure of closeness between two pdf's $p_{x}$ and $p_{y}$ is
given by the so-called {\em Kullback-Leibler divergence (KLD)},
defined as \cite{hay98} \beq
D(p_{x}\|p_{y})=E_{x}\left[\ln\frac{p_{x}(\gss{r}{}{})}{p_{y}(\gss{r}{}{})}\right]=
\int
p_{x}(\gss{r}{}{})\ln\frac{p_{x}(\gss{r}{}{})}{p_{y}(\gss{r}{}{})}d\gss{r}{}{}.
\eeq An important property of the KLD is that it is always
nonnegative and becomes zero if and only if its two arguments
coincide.\footnote{A proof follows easily by considering the
property $\ln x\leq x-1$ of the natural logarithm.} Hence,
although it is not symmetric with respect to its arguments, it is
employed as a metric for the space of pdf's.\footnote{Strictly
speaking, it is a semi-distace.} The KLD satisfies (under some
weak conditions \cite{hay98}) the following relationship, where
$p_{x},p_{y},p_{z}$ are pdf's: \beq
D(p_{x}\|p_{y})=D(p_{x}\|p_{z})+D(p_{z}\|p_{y}).
\label{eq:Pythagorean} \eeq This can be viewed as the extension
of the Pythagorean theorem for orthogonal triangles in Euclidean
spaces to the space of pdf's endowed with the metric\footnote{The
differential-geometric study of such manifolds has resulted in an
exciting new discipline, called {\em information geometry}
\cite{mr93}.} $D(\cdot\|\cdot)$ and will be seen below to provide
a means for an insightful analysis of BSS criteria.

Another property of the KLD that is going to be used in the
sequel is that it is invariant under any invertible (linear or
nonlinear) transformation, i.e., for any $\ga{g}{\cdot}$, \beq
D(p_{g(x)}\|p_{g(y)})=D(p_{x}\|p_{y}). \label{eq:D(g(y))} \eeq

The MI between $\gss{x}{}{}$ and $\gss{y}{}{}$ can be formulated
as a KLD. Indeed, using eqs.~\pref{eq:H(x)}, \pref{eq:H(x|y)},
\pref{eq:p(x,y)} in \pref{eq:I(x,y)} results in \beqa
I(\gss{x}{}{},\gss{y}{}{}) & = & \int
p_{x,y}(\gss{x}{}{},\gss{y}{}{})
\ln\frac{p_{x,y}(\gss{x}{}{},\gss{y}{}{})}{p_{x}(\gss{x}{}{})p_{y}(\gss{y}{}{})}\,
d\gss{x}{}{}d\gss{y}{}{} \nonumber \\
& = & D(p_{x,y}\|p_{x}p_{y}) \eeqa verifying the appropriateness
of the MI as a measure of statistical dependence.

The KLD formulation above makes easier to also define the mutual
information between the entries of an $N$-vector, $\gss{y}{}{}.$
It will simply equal the KLD between $p_{y}(\gss{y}{}{})$ and
$\tilde{p}_{y}(\gss{y}{}{})$ in \pref{eq:indepy}: \beq
I(\gss{y}{}{})=D(p_{y}\|\tilde{p}_{y})=\int\int \cdots \int
p_{y}(y_{1},y_{2},\ldots,y_{N})
\ln\frac{p_{y}(y_{1},y_{2},\ldots,y_{N})}{\prod_{i=1}^{N}p_{y_{i}}(y_{i})}\,dy_{1}dy_{2}\cdots
dy_{N} \eeq and vanish if and only if the components of
$\gss{y}{}{}$ are mutually independent. From the above equation
an expression of $I(\gss{y}{}{})$ in terms of the entropy of
$\gss{y}{}{}$ and the (marginal) entropies of its entries readily
results: \beq
I(\gss{y}{}{})=-H(\gss{y}{}{})+\sum_{i=1}^{N}H(y_{i}).
\label{eq:I=-H+SHi} \eeq Thus, minimizing MI amounts to making
the entropy of $\gss{y}{}{}$ as close as possible to the sum of
its marginal entropies.

{\em Negentropy}, the distance of a given pdf from the Gaussian
pdf with the same mean and covariance, can also be defined with
the aid of the KLD as \beq
J_{G}(\gss{y}{}{})\sr{\triangle}{=}D(p_{y}\|p_{y^{G}})
\label{eq:JG}, \eeq where $\gss{y}{}{G}$ denotes a Gaussian
random vector with the same mean and covariance as $\gss{y}{}{}.$

\subsection{Negative Mutual Information as an ICA Contrast}
It will be useful to write the MI of the vector $\gss{y}{}{}$ in
terms of its negentropy. This is easily done by appealing to the
Pythagorean theorem for KLD's \cite{car00b}. Just write
\pref{eq:Pythagorean} for the two triangles shown in Fig.~2.
\begin{figure}
\centerline{\psfig{figure=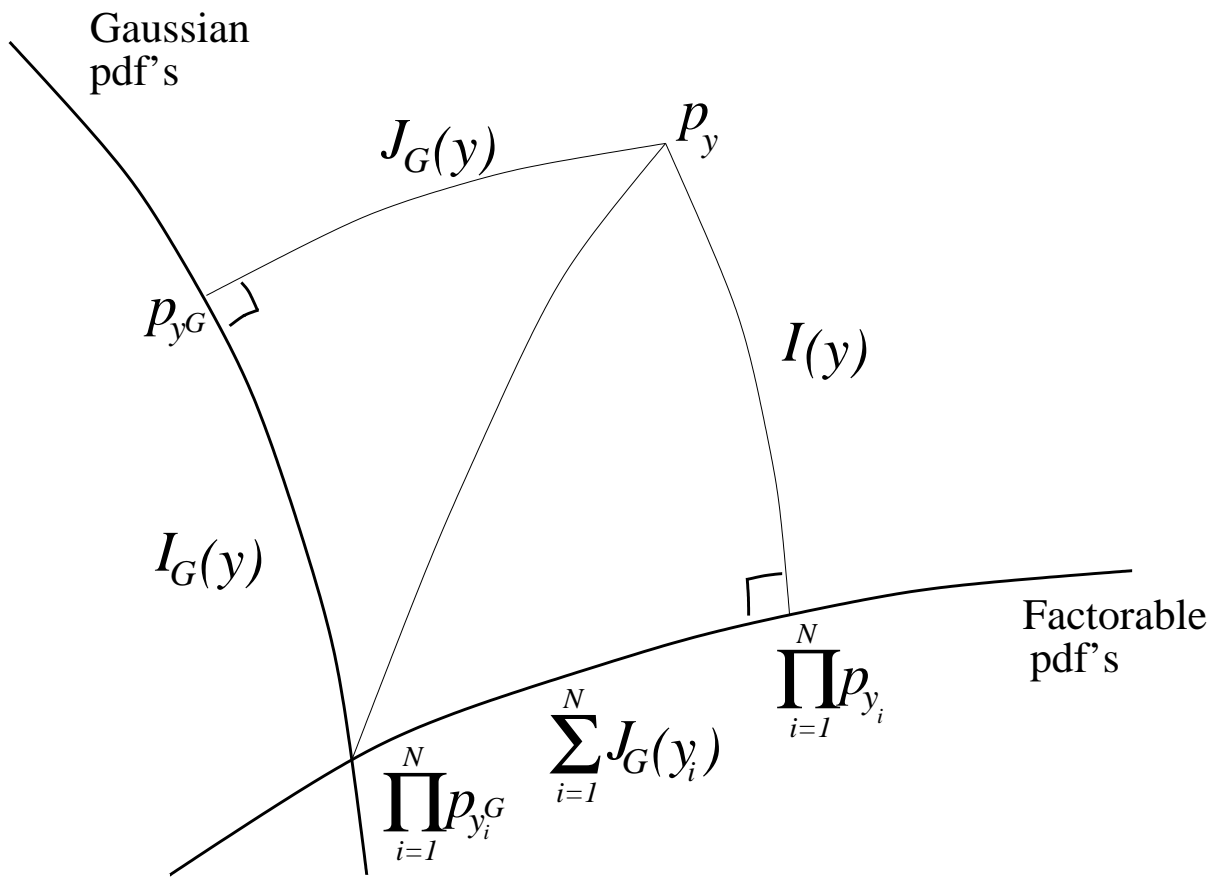,width=16 true cm}}
\caption{Information-geometric interpretation of the relation
between mutual information and negentropy \cite{car00b}.}
\end{figure}
This yields: \beqs
D\left(p_{y}\|\prod_{i=1}^{N}p_{y_{i}^{G}}\right) & = &
D(p_{y}\|p_{y^{G}})+D\left(p_{y^{G}}\|\prod_{i=1}^{N}p_{y_{i}^{G}}\right) \\
& = & J_{G}(\gss{y}{}{})+I_{G}(\gss{y}{}{}) \eeqs and \beqs
D\left(p_{y}\|\prod_{i=1}^{N}p_{y_{i}^{G}}\right) & = &
D\left(p_{y}\|\prod_{i=1}^{N}p_{y_{i}}\right)+
D\left(\prod_{i=1}^{N}p_{y_{i}}\|\prod_{i=1}^{N}p_{y_{i}^{G}}\right)
\\
& = & I(\gss{y}{}{})+\sum_{i=1}^{N}J_{G}(y_{i}), \eeqs with
$I_{G}(\gss{y}{}{})$ denoting the MI of the Gaussian version of
$\gss{y}{}{}$, $\gss{y}{}{G}.$ It then follows that\footnote{An
alternative proof for this result can be found in \cite{g99}.}
\beq
I(\gss{y}{}{})=I_{G}(\gss{y}{}{})+\left(J_{G}(\gss{y}{}{})-\sum_{i=1}^{N}J_{G}(y_{i})\right).
\label{eq:I=IG+J-SJi} \eeq The above shows $I(\gss{y}{}{})$ to
consist of two terms responsible for the redundancy within
$p_{y}$. The first term accounts for the second-order information
in the process $\gss{y}{}{}$, whereas non-Gaussian, higher-order
information is measured by the second term. The prewhitening
transformation discussed in Section~2 aims at nulling
$I_{G}(\gss{y}{}{})$. Methods based on higher-order statistics
can then be employed to minimize the second term, subject to the
constraint that the separating matrix be orthogonal. It should be
emphasized that this two-step minimization of the MI is not
necessarily optimal \cite{car00b}. Ideally, the function in
\pref{eq:I=IG+J-SJi} should be optimized as a whole. Such
approaches have only recently been reported \cite{cl96,mm98,sm98}.
It is however a common practice to first sphere the data before
going on to extract their higher-order structure. An advantage of
such an approach is that the corresponding normalization of the
data may help avoid numerical inaccuracies that could result from
the subsequent nonlinear operations \cite{lac97}. Moreover, the
fact that the separating matrix is then orthogonal may sometimes
facilitate the derivation of separation methods \cite{c94a,oit98}.
Bounds on the errors in the separation due to imperfect
standardization have been derived and reported in \cite{car94}. In
general, the standardization step poses no problem for the
separation task (provided of course one uses a sufficiently large
number of samples to estimate the autocorrelation matrix) and
two-step methods have proved their effectiveness in practice
\cite{lac97}.

It was seen above that the MI is a meaningful measure of
statistical dependence. In fact, its minimization constitutes a
valid criterion for BSS, since it can be shown that
$\psi(p_{y})=-I(\gss{y}{}{})$ is a contrast over the set of random
$N$-vectors \cite{c94a}. A real-valued function, $\Psi$, defined
on the space of $N$-variate pdf's, is said to be a {\em contrast}
if it satisfies the following requirements:
\begin{itemize}
\item[C1] $\Psi(p_{y})$ is invariant to permutations:
\[
\Psi(p_{Py})=\Psi(p_{y}) \mbox{\ for any permutation matrix\ }
\gss{P}{}{}
\]
\item[C2] $\Psi(p_{y})$ is invariant to scale changes:
\[
\Psi(p_{Dy})=\Psi(p_{y}) \mbox{\ for any diagonal matrix\ }
\gss{D}{}{}
\]
\item[C3] If $\gss{y}{}{}$ has independent components, then
\[
\Psi(p_{Ay})\leq \Psi(p_{y}), \mbox{\ for any invertible matrix\ }
\gss{A}{}{}.
\]
\end{itemize}
Requirement C3 shows that the ICA might be obtained by maximizing
a contrast of the separator's output $\gss{y}{}{}.$ The
requirements C1, C2 correspond to the permutation/scaling
indeterminacy inherent in the BSS problem, as discussed in
Section~2. However, there is one more requirement that a contrast
should satisfy so that it can be a valid BSS cost functional:
\begin{itemize}
\item[C4] The equality in C3 holds if and only if $\gss{A}{}{}$ is a
generalized permutation matrix, i.e.,
$\gss{A}{}{}=\gss{P}{}{}\gss{D}{}{}$, where $\gss{P}{}{}$ is a
permutation matrix and $\gss{D}{}{}$ is an invertible diagonal
matrix.
\end{itemize}
A contrast satisfying C4 as well is said to be {\em
discriminating.} It can be shown that the negative MI,
$\psi_{ICA}(p_{y})\sr{\triangle}{=}-I(\gss{y}{}{})$, is a
discriminating contrast over all $N$-vectors $\gss{y}{}{}$ with
{\em at most one Gaussian component} \cite{c94a}.\footnote{The
Gaussian MI, $I_{G}$, has also been shown to be a discriminating
contrast over random vector processes whose components have
different spectral densities \cite{p01}. The latter constraint is
typical in SOS-based BSS approaches as we saw in Section~3.}
Henceforth, the term contrast will be used to refer to a
discriminating contrast.

Before concluding this section, let us consider simplifying the
expression for the negative MI contrast. Taking into account the
fact that the separating matrix $\gss{G}{}{}$ is invertible, one
can express the joint pdf of $\gss{y}{}{}$ in \pref{eq:y=Gu} in
terms of that of $\gss{u}{}{}$ using the well-known formula
\cite{p84}: \beq
p_{y}(\gss{y}{}{})=\frac{p_{u}(\gss{G}{}{-1}\gss{y}{}{})}{|\det(\gss{G}{}{})|}.
\label{eq:py=pu/detG} \eeq This in turn allows the entropy of
$\gss{y}{}{}$ to be expressed in terms of that of $\gss{u}{}{}$:
\beq H(\gss{y}{}{})=H(\gss{u}{}{})+\ln|\det(\gss{G}{}{})|.
\label{eq:H(y)=H(u)+lndetG} \eeq Thus \pref{eq:I=-H+SHi} becomes
\[
I(\gss{y}{}{})=-H(\gss{u}{}{})-\ln|\det(\gss{G}{}{})|
-E_{y}\left[\ln\prod_{i=1}^{N}p_{y_{i}}(y_{i})\right]
\]
Since $H(\gss{u}{}{})$ does not depend on $\gss{G}{}{}$, the
corresponding contrast may be written as \beq
\psi_{ICA}(\gss{G}{}{})=\ln|\det(\gss{G}{}{})|+
E_{y}\left[\ln\prod_{i=1}^{N}p_{y_{i}}(y_{i})\right],
\label{eq:ICA} \eeq where, to emphasize its role in BSS, the
dependence on $\gss{G}{}{}$ is made explicit. The role of the
first term above is to ensure a valid separation solution, with
$\gss{G}{}{}$ invertible.\footnote{Recall that
$\lim_{x\rightarrow 0^{+}}\ln x=-\infty.$} Note that with
prewhitened mixture data, $\gss{G}{}{}$ is orthogonal and the
first term becomes zero. The maximization of $\psi_{ICA}$ amounts
then to minimizing the sum of the entropies of the components of
$\gss{y}{}{}.$ Since maximum entropy corresponds to Gaussianity
\cite{p84,hay98}, this is equivalent to making the $y_{i}$'s as
less Gaussian as possible, which necessarily involves statistics
of order higher than two. The central limit theorem \cite{p84}
provides another interpretation of that, since a linear mixture of
independent signals tends to be close to Gaussian. This {\em
minimum-entropy}, or {\em minimum Gaussianity} point of view has
been central in pioneering works on blind deconvolution
\cite{bgr80,d81} and will be further elaborated upon later on.

\section{A ``Universal" Criterion for BSS}
It will be of interest to examine the possible connections
between the minimum MI criterion and other proposed criteria that
are inspired from information theory and statistics. It will be
seen that subject to some constraints on the choice of the
nonlinearities involved, all these criteria are equivalent and
can thus be viewed as special cases of a single, ``universal"
criterion.
\subsection{Infomax and MaxEnt}
We have seen above that sources are completely separated when the
separator's outputs are independent or equivalently their mutual
information is zero.\footnote{Provided, of course, that at most
one of the sources is Gaussian.} An alternative way of looking at
the separation goal is by formulating it as a maximal information
transfer from the mixture, $\gss{u}{}{}$, to the source
estimates, $\gss{y}{}{}$ \cite{b00}. This criterion, similar to
the principle of information rate and capacity in the Shannon
theory of communication, though less evidently pertinent to the
BSS task, will be seen to be closely related to that of minimum
MI between the entries of $\gss{y}{}{}.$

Consider the MI between $\gss{u}{}{}$ and $\gss{y}{}{}$, namely:
\[
I(\gss{y}{}{},\gss{u}{}{})=H(\gss{y}{}{})-H(\gss{y}{}{}|\gss{u}{}{}).
\]
Since the transfer from $\gss{u}{}{}$ to $\gss{y}{}{}$ is
deterministic, knowledge of the first completely determines the
latter, hence the second term above is null. Therefore, in this
context: \beq I(\gss{y}{}{},\gss{u}{}{})=H(\gss{y}{}{}).
\label{eq:I(y,u)=H(y)} \eeq Since for an unrestricted
$\gss{y}{}{}$ its entropy has no bound, we assume an entrywise
nonlinearity at the output of the separating system $\gss{G}{}{}$
(see Fig.~1), i.e., \beq \ga{y}{n}=\ga{g}{\gss{G}{}{}\ga{u}{n}}
\eeq
where
\[
\ga{g}{x}\sr{\triangle}{=}\bmx{cccc} g_{1}(x_{1}) & g_{2}(x_{2})
& \cdots & g_{N}(x_{N}) \emx^{T}
\]
and the functions $g_{i}$ are monotonically increasing with
$g_{i}(-\infty)=0$ and $g_{i}(\infty)=1$. These are the common
assumptions for the nonlinearities in artificial neural networks
(ANN) \cite{hay98}. Then the entropy of $\gss{y}{}{}$ is maximized
when $p_{y}(\gss{y}{}{})$ equals $U_{N}(\gss{y}{}{})$, the
uniform pdf in $[0,1]^{N}.$ Note that in that case, the
components of $\gss{y}{}{}$ are statistically independent and
uniformly distributed in $[0,1]$. If $g_{i}$ is chosen to be
equal to the cumulative distribution function (cdf) of the source
$i$, $p_{y_{i}}(y_{i})$ can only be uniform in $[0,1]$ when
$y_{i}$ is equal to $a_{i}$ or to another source having the same
pdf. Therefore the maximization of the entropy of $\gss{y}{}{}$
yields a meaningful criterion for BSS, called {\em MaxEnt}
\cite{bs95}. In view of \pref{eq:I(y,u)=H(y)}, this criterion is
equivalent to the maximization of the MI
$I(\gss{y}{}{},\gss{u}{}{})$, referred to as {\em Infomax}
\cite{bs95,b00}. The above source-output pdf-matching point of
view was also considered in early studies of blind equalization
\cite{bgr80} and will be seen to underly the maximum likelihood
approach as well.

The MaxEnt and Infomax criteria can be seen to be equivalent,
under some conditions, with the minimum MI criterion discussed in
Section~4. This equivalence has been argued upon in \cite{bs95}
considering the functions $g_{i}$ as being given by the source
pdf's, as explained above. However, a more rigorous study was
reported in \cite{ya97} and \cite{od98}. Here a brief explanation
of this relation will be provided stating more subtle results
without proof. Denote by $\gss{x}{}{}$ the linear part of the
separating system (see Fig.~1), i.e., \beq
\ga{x}{n}=\gss{G}{}{}\ga{u}{n}. \eeq Then, since $\ga{g}{\cdot}$
is invertible it holds that \cite{p84}
\[
p_{y}(\gss{y}{}{})=
\frac{p_{x}(\gss{x}{}{})}{\left|\prod_{i=1}^{N}g_{i}^{\prime}(x_{i})\right|},
\]
with $g_{i}^{\prime}$ denoting the first derivative of $g_{i}.$
The entropy of $\gss{y}{}{}$ can thus be written
as\footnote{Recall that $g_{i}$ is nonnegative.} \beq
H(\gss{y}{}{})=H(\gss{u}{}{})+\ln|\det(\gss{G}{}{})|+
E_{x}\left[\ln\prod_{i=1}^{N}g_{i}^{\prime}(x_{i})\right]. \eeq
Hence the Infomax (IM), or equivalently MaxEnt, criterion reads
as: \beq \max_{G}\left(\psi_{IM}(\gss{G}{}{})\sr{\triangle}{=}
\ln|\det(\gss{G}{}{})|+E_{x}\left[\ln\prod_{i=1}^{N}g_{i}^{\prime}(x_{i})\right]\right).
\label{eq:IM} \eeq Comparing eqs.~\pref{eq:ICA} and \pref{eq:IM},
one can see that these two criteria are equivalent if $g_{i}$'s
coincide with source cdf's.\footnote{Note that in both cases, the
expectation in the last term is with respect to the linear part
of the demixing system.} Eq.~\pref{eq:I=IG+J-SJi} expresses the
MI between the components of $\gss{y}{}{}$ in terms of its joint
and marginal negentropies. It has been demonstrated that Infomax
is also equivalent with negentropy maximization, under the same
conditions \cite{lgbs00}. In fact, since it is the stability of
the desired stationary point $\gss{G}{}{}=\gss{H}{}{-1}$ that
counts, it can be seen that a mismatch between the $g_{i}$'s and
the source probability laws can be tolerated. This is stated in
\cite{od98} as a requirement for {\em sufficiently rich
parameterization} of $g_{i}$. The problem of correctly choosing
the nonlinearities $g_{i}$ will be discussed in more detail later
on when dealing with the corresponding algorithms.

\subsection{Maximum Likelihood and Bayesian Approaches}
Let us continue to consider the demixing model
$\gss{y}{}{}=\ga{g}{\gss{x}{}{}}$ where the $g_{i}$'s are always
taking values in $[0,1].$ Then the entropy of $\gss{y}{}{}$ can
be written as the negative distance of its pdf from the uniform
one: \beqa H(\gss{y}{}{}) & = & -\int
p_{y}(\gss{y}{}{})\ln\left[\frac{p_{y}(\gss{y}{}{})}{\prod_{i=1}^{N}U(y_{i})}\right]
d\gss{y}{}{} \nonumber \\
& = & -D(p_{y}\|U_{N}), \label{eq:H=D(p||U)} \eeqa where
$U(y_{i})$ is the uniform pdf in $[0,1]$ and
$U_{N}(\gss{y}{}{})=\prod_{i=1}^{N}U(y_{i}).$ In order to
formulate a {\em Maximum Likelihood (ML)} approach to the
corresponding BSS problem, we need a model for the observed data
generation including a hypothesis for the source statistics.
According to \pref{eq:u=Ha} this will be written as
\[
\gss{u}{}{}=\gss{\Theta}{}{}\gss{\tilde{a}}{}{}
\]
where ideally $\gss{\Theta}{}{}=\gss{H}{}{}$ and
$\gss{\tilde{a}}{}{}=\gss{a}{}{}.$ Parameterize the pdf of
$\gss{u}{}{}$ as $p_{u}(\gss{u}{}{}|\gss{\Theta}{}{}).$ Although
this pdf depends on both the system parameter and our
hypothesized probability model for the sources,
$\gss{\tilde{a}}{}{}$, the latter will be omitted from this
notation since it will be with respect to $\gss{\Theta}{}{}$ that
the criterion will be optimized. The idea in the ML principle is
to find, among a set of choices for $\gss{\Theta}{}{}$, that
which maximizes the pdf of the data conditioned on the model
parameter.

Take $T$ independent realizations of $\gss{u}{}{}$, say,
$\ga{u}{1},\ldots,\ga{u}{T}.$ Then the likelihood that these
samples are drawn with a pdf
$p_{u}(\gss{u}{}{}|\gss{\Theta}{}{})$ is given by
$\prod_{t=1}^{T}p_{u}(\ga{u}{t}|\gss{\Theta}{}{}).$ The
normalized log-likelihood will be
\[
L_{T}(\gss{\Theta}{}{})=\frac{1}{T}\ln\prod_{t=1}^{T}p_{u}(\ga{u}{t}|\gss{\Theta}{}{})
=\frac{1}{T}\sum_{t=1}^{T}\ln p_{u}(\ga{u}{t}|\gss{\Theta}{}{}).
\]
As in \pref{eq:py=pu/detG}, we have \beqs
p_{u}(\ga{u}{t}|\gss{\Theta}{}{}) & = &
\frac{p_{\tilde{a}}(\gss{\Theta}{}{-1}\ga{u}{t})}{|\det(\gss{\Theta}{}{})|}
\\
& = & p_{\tilde{a}}(\ga{x}{t})|\det(\gss{G}{}{})|
 \eeqs
where we recall that $\gss{x}{}{}=\gss{G}{}{}\gss{u}{}{}$ and we
set $\gss{G}{}{}=\gss{\Theta}{}{-1}.$ Hence,
\[
L_{T}(\gss{\Theta}{}{})=\frac{1}{T}\sum_{t=1}^{T}\ln
p_{\tilde{a}}(\ga{x}{t})+\ln|\det(\gss{G}{}{})|
\]
and from the law of large numbers \cite{p84}:
\[
L_{T}(\gss{\Theta}{}{})\sr{T\rightarrow\infty}{\longrightarrow}L(\gss{G}{}{})
\sr{\triangle}{=}\int p_{x}(\gss{x}{}{}|\gss{G}{}{})\ln
p_{\tilde{a}}(\gss{x}{}{})d\gss{x}{}{}+\ln|\det(\gss{G}{}{})|.
\]
Using the KLD formulation, the latter is written as \beqs
L(\gss{G}{}{}) & = &
-D(p_{x}(\gss{x}{}{}|\gss{G}{}{})\|p_{\tilde{a}}(\gss{x}{}{}))-H(\gss{x}{}{}|\gss{G}{}{})
+\ln|\det(\gss{G}{}{})| \\
& = &
-D(p_{x}(\gss{x}{}{}|\gss{G}{}{})\|p_{\tilde{a}}(\gss{x}{}{}))-H(\gss{u}{}{}),
\eeqs where eq.~\pref{eq:H(y)=H(u)+lndetG} was used. The ML
criterion is then expressed in the form: \beq
\max_{G}\left(\psi_{ML}(\gss{G}{}{})\sr{\triangle}{=}
-D(p_{x}(\gss{x}{}{}|\gss{G}{}{})\|p_{\tilde{a}}(\gss{x}{}{}))\right).
\label{eq:ML1} \eeq That is, it aims at matching the pdf's of the
source estimates with the hypothesized source pdf's.

For the sake of clarity of presentation, we shall abuse the
notation and write $D(\gss{x}{}{}\|\gss{y}{}{})$ for the KLD
between the pdf's of the random vectors $\gss{x}{}{}$ and
$\gss{y}{}{}.$ Then the ML cost functional is written in the form:
\beq
\psi_{ML}(\gss{G}{}{})=-D(\gss{G}{}{}\gss{H}{}{}\gss{a}{}{}\|\gss{\tilde{a}}{}{}).
\label{eq:ML} \eeq Now assume that the $g_{i}$'s are the cdf's
corresponding to the hypothesized probability law for the
sources, $\gss{\tilde{a}}{}{}.$ In that case, the vector
$\gss{z}{}{}=\ga{g}{\gss{\tilde{a}}{}{}}$ is uniformly
distributed in $[0,1]^{N}$ and, in view of \pref{eq:H=D(p||U)},
$\psi_{IM}(\gss{G}{}{})$ can be written as: \beqa
\psi_{IM}(\gss{G}{}{}) & = & H(\gss{y}{}{}) \nonumber \\
& = & -D(\gss{y}{}{}\|\gss{z}{}{}) \nonumber \\
& = &
-D(\ga{g}{\gss{G}{}{}\gss{H}{}{}\gss{a}{}{}}\|\ga{g}{\gss{\tilde{a}}{}{}{}})
\nonumber \\
& = & -D(\gss{G}{}{}\gss{H}{}{}\gss{a}{}{}\|\gss{\tilde{a}}{}{})
\label{eq:IM2} \eeqa where the invariance of the KLD under an
invertible transformation was exploited. Comparing
eqs.~\pref{eq:ML} and \pref{eq:IM2} the equivalence of ML with
the IM criterion is deduced \cite{car97}. Clearly, if the source
pdf's are exactly known, i.e., $\gss{\tilde{a}}{}{}=\gss{a}{}{},$
both cost functions above are maximized (become zero) when
$\gss{G}{}{}=\gss{H}{}{-1}$ or equivalently
$\gss{\Theta}{}{}=\gss{H}{}{},$ that is, for the perfect
separation solution. Nevertheless, as mentioned earlier, there is
some tolerance in the mismatch of the source probability model,
quantified by the stability conditions stated in Section~6.

An insightful interpretation of the ML criterion can be obtained
by making use in \pref{eq:ML} of the Pythagorean theorem for the
KLD's (eq.~\pref{eq:Pythagorean}). If the components of
$\gss{\tilde{a}}{}{}$ are independent and $\gss{\tilde{y}}{}{}$
denotes the vector with independent components distributed
according to the marginal distributions of $\gss{y}{}{}$, then
\beq D(\gss{y}{}{}\|\gss{\tilde{a}}{}{})=
D(\gss{y}{}{}\|\gss{\tilde{y}}{}{})+D(\gss{\tilde{y}}{}{}\|\gss{\tilde{a}}{}{}).
\eeq Eq.~\pref{eq:ML} then yields: \beqa -\psi_{ML}(\gss{G}{}{})
& = &
D(\gss{y}{}{}\|\gss{\tilde{y}}{}{})+D(\gss{\tilde{y}}{}{}\|\gss{\tilde{a}}{}{})
\nonumber \\
& = &
-\psi_{ICA}(\gss{G}{}{})+\sum_{i=1}^{N}D(\tilde{y}_{i}\|\tilde{a}_{i})
\eeqa which leads to the following interpretation of ML
\cite{car98a}: \beq \left(\begin{array}{c} {\rm Total} \\ {\rm
mismatch}
\end{array}\right)=
\left(\begin{array}{c} {\rm Deviation\ from} \\
{\rm independence}\end{array}\right)+ \left(\begin{array}{c} {\rm
Marginal} \\ {\rm mismatch}
\end{array}\right) \eeq
In fact, the above relationship shows that the ICA criterion
results from optimizing the KLD of \pref{eq:ML} with respect to
both $\gss{G}{}{}$ {\em and} the source probability model. In
other words,
\[
\psi_{ICA}(\gss{G}{}{})=\max_{\tilde{a}}\psi_{ML}(\gss{G}{}{}).
\]
This shows $I(\gss{y}{}{})$ to be the quantitative measure of
dependence associated with the ML principle \cite{car98a}.

A criterion which is equivalent to the above results also via the
Bayesian approach \cite{k99,md99}. By Bayes' theorem \cite{p84}
one may write the a-posteriori pdf of the model
$(\gss{\Theta}{}{},\gss{a}{}{})$ in terms of the likelihood.
Namely:
\[
p(\gss{\Theta}{}{},\gss{a}{}{}|\gss{u}{}{},{\cal I})=
\frac{\overbrace{p(\gss{u}{}{}|\gss{\Theta}{}{},\gss{a}{}{},{\cal
}I)}^{p_{u}(u|\Theta)}p(\gss{\Theta}{}{},\gss{a}{}{}|{\cal
}I)}{p(\gss{u}{}{}|{\cal I})}
\]
where ${\cal I}$ is used to denote any possible additional prior
information on the BSS setup. One can simplify the above
expression as
\[
p(\gss{\Theta}{}{},\gss{a}{}{}|\gss{u}{}{},{\cal I}) \propto
p(\gss{u}{}{}|\gss{\Theta}{}{},\gss{a}{}{},{\cal
I})p(\gss{\Theta}{}{},\gss{a}{}{}|{\cal I})
\]
where the prior probability of the data was incorporated in the
proportionality constant since it does not depend explicitly on
the model. Since the mixing matrix is in general independent of
the sources, we write
\[
p(\gss{\Theta}{}{},\gss{a}{}{}|{\cal I})=p(\gss{\Theta}{}{}|{\cal
I})p(\gss{a}{}{}|{\cal I}).
\]
Finally, since it is the determination of the demixing matrix
$\gss{\Theta}{}{-1}$ that is of interest, we consider the
following a-posteriori model:
\[
p(\gss{\Theta}{}{}|\gss{u}{}{},{\cal I}) \propto
p(\gss{\Theta}{}{}|{\cal I})\int
p(\gss{u}{}{}|\gss{\Theta}{}{},\gss{a}{}{},{\cal
I})p(\gss{a}{}{}|{\cal I})d\gss{a}{}{}
\]
or its logarithm: \beq \ln p(\gss{\Theta}{}{}|\gss{u}{}{},{\cal
I})= \ln p(\gss{\Theta}{}{}|{\cal I})+\ln\int
p(\gss{u}{}{}|\gss{\Theta}{}{},\gss{a}{}{},{\cal
I})p(\gss{a}{}{}|{\cal I})d\gss{a}{}{} \label{eq:lnMAP}\eeq
where
a constant was omitted as it has no effect on the subsequent
optimization.

Consider the {\em Maximum A-Posteriori probability (MAP)}
criterion: \beq
\max_{\Theta}\left(\psi_{MAP}(\gss{\Theta}{}{})\sr{\triangle}{=}E[\ln
p(\gss{\Theta}{}{}|\gss{u}{}{},{\cal I})]\right)
\label{eq:MAP1}\eeq The essence of the Bayesian approach is to
make an inference on the unknown model parameters on the basis of
all information available \cite{k99}. The information that
$\gss{u}{}{}=\gss{\Theta}{}{}\gss{a}{}{}$ is exploited by setting
the likelihood equal to:
\[
p(\gss{u}{}{}|\gss{\Theta}{}{},\gss{a}{}{},{\cal I})=
\prod_{i=1}^{N}\delta\left(u_{i}-\sum_{j=1}^{N}\Theta_{ij}a_{j}\right)
\]
with $\delta(\cdot)$ denoting the Dirac function. The knowledge
concerning the independence of the sources is made use of by using
\[
p(\gss{a}{}{}|{\cal I})=\prod_{i=1}^{N}p_{a_{i}}(a_{i}|{\cal I}).
\]
Since no specific knowledge is available for the mixing matrix \,
apart from its dimensions and its nonsingularity,
$p(\gss{\Theta}{}{}|{\cal I})$ is set to a constant\footnote{With
a finite extent, of course.} and is thus omitted from the MAP
criterion. In summary, \beqs \int
p(\gss{u}{}{}|\gss{\Theta}{}{},\gss{a}{}{},{\cal
I})p(\gss{a}{}{}|{\cal I})d\gss{a}{}{} & = & \int\int\cdots\int
\prod_{i=1}^{N}\delta\left(u_{i}-\sum_{j=1}^{N}\Theta_{ij}a_{j}\right)
\prod_{i=1}^{N}p_{a_{i}}(a_{i}|{\cal I})da_{1}da_{2}\cdots da_{N}
\\
& = &
\det(\gss{G}{}{})\prod_{i=1}^{N}p_{a_{i}}\left(\sum_{j=1}^{N}G_{ij}u_{j}\right)
\eeqs with $\gss{G}{}{}=\gss{\Theta}{}{-1}.$ The MAP cost
functional then takes the form: \beq
\psi_{MAP}(\gss{G}{}{})=\ln|\det(\gss{G}{}{})|+E\left[\ln\prod_{i=1}^{N}p_{a_{i}}(x_{i})\right].
\label{eq:MAP} \eeq A comparison of the latter expression with
that in \pref{eq:IM} demonstrates the equivalence of the Bayesian
approach with the Infomax, provided that the nonlinearities
$g_{i}$ match the source cdf's.

\subsection{A Universal Criterion}
In view of the above results, one can define a general BSS
criterion in terms of the maximization with respect to
$\gss{G}{}{}$ of the expectation of the functional \beq
J(\gss{G}{}{})=\ln|\det(\gss{G}{}{})|+\ln\prod_{i=1}^{N}\phi_{i}(y_{i})
\label{eq:uniBSS} \eeq where $\gss{y}{}{}=\gss{G}{}{}\gss{u}{}{}$.
The functions $\phi_{i}$ are ideally the pdf's of the source
signals. Since these are in general unknown, some approximations
have to be made. A short discussion of such approaches, mainly
via pdf expansions, is given here. More will be said in the next
section where the role of the nonlinearities $\phi_{i}$ in the
stability of the separator will be reviewed.
\subsubsection{Approximations}
The criteria discussed above require a knowledge of pdf's
connected to the separator's outputs. The marginal pdf's of
$\gss{y}{}{}$ are needed for example in $\psi_{ICA}.$ An approach
towards overcoming the lack of this knowledge, which has been
proved particularly effective, it to employ truncated expansions
of the unknown pdf's and thus reduce the unknowns to a limited
set of cumulants, computable from the system outputs. To give an
idea of this approach, let us recall the Edgeworth expansion of a
pdf $p(y)$ of zero mean and unit variance, around a standard
normal pdf, $p_{G}(y)$ \cite{lac97}: \beqa p(y) & = &
p_{G}(y)\left[1+\frac{1}{3!}c_{3}(y)h_{3}(y)+\frac{1}{4!}c_{4}(y)h_{4}(y)+
\frac{10}{6!}c_{3}^{2}(y)h_{6}(y)\right. \nonumber \\ & & \mbox{\
\ \ \ \ \ \ \ \ \ \ \ \ \ \ \ \ }+
\left.\frac{1}{5!}c_{5}(y)h_{5}(y)+\frac{35}{7!}c_{3}(y)c_{4}(y)h_{7}(y)+\cdots\right]
\label{eq:Edgeworth} \eeqa where $c_{k}(y)$ is the $k$th-order
cumulant of $p(y)$ and $h_{k}(y)$ is the Hermite polynomial of
degree $k$, defined by \cite{hay98}:
\[
\frac{d^{k}p_{G}(y)}{dy^{k}}=(-1)^{k}p_{G}(y)h_{k}(y)
\]
or equivalently by the recursion
\beqs h_{0}(y) & = & 1 \\
h_{1}(y) & = & y \\
h_{k+1}(y) & = & yh_{k}(y)-h_{k}^{\prime}(y). \eeqs Keeping only
the first few terms in the sum \pref{eq:Edgeworth} one comes up
with a polynomial approximation of $p_{y_{i}}(y_{i})$ in
\pref{eq:ICA}. Using a Taylor approximation for the logarithm
function finally leads to an approximative expression of the cost
functional in terms of some cumulants of the $y_{i}$'s. In the
case that a prewhitening has been performed on the data, it can be
shown that, when data are symmetrically distributed (so that
cumulants of odd order are null), a good approximation is
provided by \cite{c94a}: \beq \psi_{ICA}(\gss{G}{}{})\approx
\psi_{4}(\gss{G}{}{})\sr{\triangle}{=}\sum_{i=1}^{N}c_{4}^{2}(y_{i})
\label{eq:cum4}\eeq where it should be kept in mind that
$\gss{G}{}{}$ is constrained to be orthogonal. As shown in
\cite{c94a}, $\psi_{4}$ is a (discriminating) contrast over the
random $N$-vectors with at most one component of null 4th-order
cumulant. Note that this constraint is stronger than that applied
to $\psi_{ICA}$, since a signal has to have {\em all} its
high-order cumulants equal to zero to be Gaussian. The same cost
function has been arrived at via a Gram-Charlier truncated
expansion for the ML criterion \cite{lac97}. It is interesting to
note that, due to the orthogonality hypothesis,
$\max_{G}\psi_{4}(\gss{G}{}{})$ is equivalent to forcing the
cross-cumulants of $\gss{y}{}{}$ to zero, thus rendering its
components mutually independent.\footnote{It has been recently
shown, however, that such a separation condition may be greatly
simplified to the requirement of zeroing a much smaller set of
cumulants \cite{np97}.} We shall have more to say about that in
the context of algebraic BSS approaches.

Polynomial approximations of the above type may not be
sufficiently accurate in practice. This is due to the fact that
finite-sample cumulant estimates are highly sensitive to
outliers: a few, possibly erroneous observations with large
values may prove to be decisive for the accuracy of the resulting
estimate. To cope with this problem, alternative methods for
estimating the entropy have been devised, which perform better
than the cumulant-based estimates and at a comparable
computational cost. These include \cite{h98,od00}.

\subsubsection{Nonlinear PCA}
PCA aims at best approximating, in the least-squares (LS) sense,
the given data by a {\em linear} projection on a set of orthogonal
vectors. As explained in Section~3, this is not sufficient to
extract from the data its higher-order structure. A
generalization of PCA, involving a {\em nonlinear} projection, has
been proposed and demonstrated to be effective in the BSS
problem. This so-called {\em nonlinear PCA} approach can be
described in terms of the minimization of the cost functional
\cite{kpo98} \beq \psi_{nPCA}(\gss{G}{}{})=
E\left[\|\gss{u}{}{}-\gss{G}{}{T}\ga{g}{\gss{G}{}{}\gss{u}{}{}}\|^{2}\right].
\label{eq:nPCA} \eeq Assuming data prewhitening, $\gss{G}{}{}$ is
orthogonal and the above can be rewritten as: \beqa
\psi_{nPCA}(\gss{G}{}{}) & = &
E\left[\|\gss{G}{}{}\gss{u}{}{}-\ga{g}{\gss{G}{}{}\gss{u}{}{}}\|^{2}\right]
\nonumber \\
& = & E\left[\|\gss{y}{}{}-\ga{g}{\gss{y}{}{}}\|^{2}\right]
\nonumber \\
& = & \sum_{i=1}^{N}E[|y_{i}-g_{i}(y_{i})|^{2}].
\label{eq:Bussgang} \eeqa The latter form of the nonlinear PCA
cost reveals its close connection with the so-called Bussgang
approach to blind equalization \cite{b94}, where a LS cost is
adopted with the nonlinearly transformed output, $g(y)$, playing
the role of the unavailable desired response. Moreover, for
appropriate choices of $\ga{g}{\cdot}$, the above criterion can be
shown to be well approximated by the maximization of the sum of
the absolute values of the autocumulants of $\gss{y}{}{}$
\cite{l98,g99,lgbs00}, which is in turn equivalent to the
criterion defined by \pref{eq:cum4}.
\section{Adaptive Algorithms}
This section will consider the design of adaptive stochastic
gradient algorithms for the criterion \pref{eq:uniBSS}. Taking
the derivative with respect to $\gss{G}{}{}$ of the functional in
\pref{eq:uniBSS} yields: \beq \nabla
J(\gss{G}{}{})=\gss{G}{}{-T}-\ga{f}{\gss{y}{}{}}\gss{u}{}{T} \eeq
where the vector function $\ga{f}{\cdot}$ is defined as: \beq
\ga{f}{\gss{y}{}{}}=\bmx{cccc}
-\frac{\phi_{1}^{\prime}(y_{1})}{\phi_{1}(y_{1})} &
-\frac{\phi_{2}^{\prime}(y_{2})}{\phi_{2}(y_{2})} & \cdots &
-\frac{\phi_{N}^{\prime}(y_{N})}{\phi_{N}(y_{N})} \emx^{T}. \eeq
The derivatives of the logarithms of $\phi_{i}$,
$\frac{\phi_{i}^{\prime}(y_{i})}{\phi_{i}(y_{i})}$, are known as
the {\em score functions} \cite{car98a}. The gradient ascent
algorithm will then be of the form: \beq
\ga{G}{k+1}=\ga{G}{k}+\Delta\ga{G}{k} \label{eq:ascent}\eeq where
\beq \Delta\ga{G}{k}\propto
\gssa{G}{}{-T}{k}-\ga{f}{\ga{y}{k}}\gssa{u}{}{T}{k}.
\label{eq:Infomax} \eeq Using the logistic sigmoid for the
functions $\phi_{i}$,
\[
\phi_{i}(y_{i})=\frac{1}{1+e^{-y_{i}}},
\]
the above recursion yields the Infomax algorithm, first proposed
in \cite{bs95}.

The connection of the algorithm \pref{eq:Infomax} with the
nonlinear PCA rule was studied in \cite{kpo98,pk99}. The latter
is described by: \beq \Delta \gssa{G}{}{T}{k}\propto
\left[\ga{u}{k}-\gssa{G}{}{T}{k}\ga{f}{\ga{y}{k}}\right]\ga{f}{\gssa{y}{}{T}{k}}
\label{eq:nlPCA} \eeq where $\gss{G}{}{}$ is constrained to be
orthogonal (see Section~5.3.2). With the approximation
\[ E[\ga{f}{\ga{y}{k}}\ga{f}{\gssa{y}{}{T}{k}}]\approx \gss{I}{}{}, \]
valid for almost independent $y_{i}$'s, \pref{eq:nlPCA} becomes
\[
\Delta\gssa{G}{}{T}{k}\propto
-\left[\gssa{G}{}{T}{k}-\ga{u}{k}\ga{f}{\gssa{y}{}{T}{k}}\right]
\]
or equivalently \beq \Delta\ga{G}{k}\propto
-\left[\ga{G}{k}-\ga{f}{\ga{y}{k}}\gssa{u}{}{T}{k}\right].
\label{eq:anlPCA}\eeq In view of the orthogonality of
$\gss{G}{}{}$, \pref{eq:anlPCA} coincides with \pref{eq:Infomax}
but with a reversed sign.

\subsection{Equivariance -- Relative/Natural Gradient}
In addition to the need for an inversion of an $N\times N$ matrix
at each iteration, the algorithm of \pref{eq:Infomax} has the
disadvantage of not guaranteeing the invertibility of this matrix
in the process of adaptation. Moreover, it can readily be seen
that the separation performance obtained depends on the mixing
matrix in question. One would like to have an algorithm able of
yielding a uniform performance, independent of the ill- or
well-conditioning of $\gss{H}{}{}.$ This feature can be obtained
via a so-called {\em equivariant} estimator of the mixing matrix,
namely one that produces estimates that, under data
transformation, are transformed accordingly
\cite{cl96,car98a,car00a}: \beq \hga{\cal
H}{\gss{G}{}{}\gss{u}{}{}}=\gss{G}{}{}\hga{\cal H}{\gss{u}{}{}}
\label{eq:equiv} \eeq for any nonsingular $N\times N$ matrix
$\gss{G}{}{}.$ Then the source estimate will be given by: \beqs
\hgss{a}{}{} & = & (\hga{\cal H}{\gss{u}{}{}})^{-1}\gss{u}{}{} \\
& = & (\hga{\cal
H}{\gss{H}{}{}\gss{a}{}{}})^{-1}\gss{H}{}{}\gss{a}{}{} \\
& = & (\gss{H}{}{}\hga{\cal
H}{\gss{a}{}{}})^{-1}\gss{H}{}{}\gss{a}{}{} \\
& = & (\hga{\cal H}{\gss{a}{}{}})^{-1}\gss{a}{}{}, \eeqs which is
seen not to depend on the mixing matrix.

The algorithm \pref{eq:Infomax} can be transformed so as to be
equivariant by postmultiplying the right-hand side with
$\gssa{G}{}{T}{k}\ga{G}{k}$. Since this matrix is positive
definite, this operation will not affect the categorization and
stability of stationary points. The new recursion is
\pref{eq:ascent} with: \beq \Delta\ga{G}{k} \propto
\left(\gss{I}{}{}-\ga{f}{\ga{y}{k}}\gssa{y}{}{T}{k}\right)\ga{G}{k}.
\label{eq:equivInfomax}\eeq It can be shown that this update
preserves the invertibility of $\gss{G}{}{}$ \cite{ac98}. The
fact that it also enjoys the equivariance property is made easily
clear by looking at the corresponding update for the {\em global}
or {\em combined} system \beq \gss{S}{}{}=\gss{G}{}{}\gss{H}{}{}.
\eeq Postmultiplying \pref{eq:equivInfomax} with $\gss{H}{}{}$,
the following update results that depends only on $\gss{S}{}{}$:
\beq \Delta\ga{S}{k}\propto
\left(\gss{I}{}{}-\ga{f}{\ga{y}{k}}\gssa{y}{}{T}{k}\right)\ga{S}{k}.
\label{eq:equivS} \eeq The equivariance stems from the fact that
this process depends on $\gss{H}{}{}$ only through its
initialization $\ga{S}{0}=\ga{G}{0}\gss{H}{}{}$, which may remain
unchanged when $\gss{H}{}{}$ is modified simply by appropriately
changing the initialization of $\gss{G}{}{}$
\cite{cl96}.\footnote{Of course, this argument does not apply in
the general case where $\gss{H}{}{}$ is not invertible; see
Section~9.}

The above transformation can be interpreted with the aid of the
so-called {\em relative gradient} \cite{cl96,car00a}, defined for
a function $\Phi$ as: \beq
\tilde{\nabla}\Phi(\gss{G}{}{})\sr{\triangle}{=}\left.\frac{\partial
\Phi((\gss{I}{}{}+\gss{\cal E}{}{})\gss{G}{}{})}{\partial
\gss{\cal E}{}{}}\right|_{{\cal E}=0}. \label{eq:relg} \eeq That
is, $\tilde{\nabla}\Phi(\gss{G}{}{}$ measures the rate of
variation of $\Phi$ with respect to {\em relative} changes at
$\gss{G}{}{}$, hence its name. This complies well with the {\em
serial update} in \pref{eq:equivInfomax} \cite{cl96}. The role of
the relative gradient in the transformation above is seen in its
relation with the gradient of $\Phi$, as it follows directly from
its definition \cite{cl96,car98a,car00a}: \beq
\tilde{\nabla}\Phi(\gss{G}{}{})=\nabla
\Phi(\gss{G}{}{})\gss{G}{}{T}.\eeq

A closely related notion is that of the {\em natural gradient}
\cite{a98,ac98,da00}.\footnote{See \cite{car98c} for a rigorous
treatment of the relation between these two gradient
definitions.} It was conceived to address the fact that the
simple gradient does not always define the steepest ascent
direction of a function. In fact, the magnitude of the gradient
varies in general with the direction from a (local) maximum,
hence the convergence of a gradient ascent scheme may be quite
slow for some initializations. A sophisticated schedule for
varying the step size would be required, though it would be
impossible to be applied in practice due to lack of knowledge of
the function shape. The problem lies in the fact that in some
applications, including BSS, the parameter space is not
Euclidean, hence a different distance measure should be employed
to derive the steepest ascent direction. The natural gradient
provides the solution for a general, Riemannean manifold, as the
multiplicative group of invertible matrices $\gss{G}{}{}$ is. It
is shown to mitigate the anisotropy problem mentioned above and
moreover, when used in gradient updates, provides Newton-like
behavior even when the function is not quadratic around its
maximum \cite{cl96,a98,ac98,da00}. It has also been shown to be
Fisher-efficient \cite{a98}. Its ``disadvantage" is that one has
to know the structure of the parameter space to properly define
it \cite{da00}. However, this is not a problem in the BSS context
where the natural gradient has given rise to a great many
efficient algorithms for both static and dynamic mixtures
\cite{ac98}.

\subsection{Nonlinear Anti-Hebbian Learning}
It will be instructive to view the above algorithms as being
instances of the principle of nonlinear anti-Hebbian learning.
The linear anti-Hebbian rule for adjusting the synaptic strength
between two neural cells with activities $y_{i}$ and $y_{j}$ is
generally given as follows \cite{hay98,g99}: \beq \Delta
G_{ij}\propto -y_{i}y_{j}. \label{eq:lHebb} \eeq If $y_{i},y_{j}$
are positively (negatively) correlated, this rule will build a
negative (positive) weight between them that makes simultaneous
firing more difficult and therefore eliminates their correlation.
A two-neuron ANN model for this is shown in Fig.~3.
\begin{figure}
\centerline{\psfig{figure=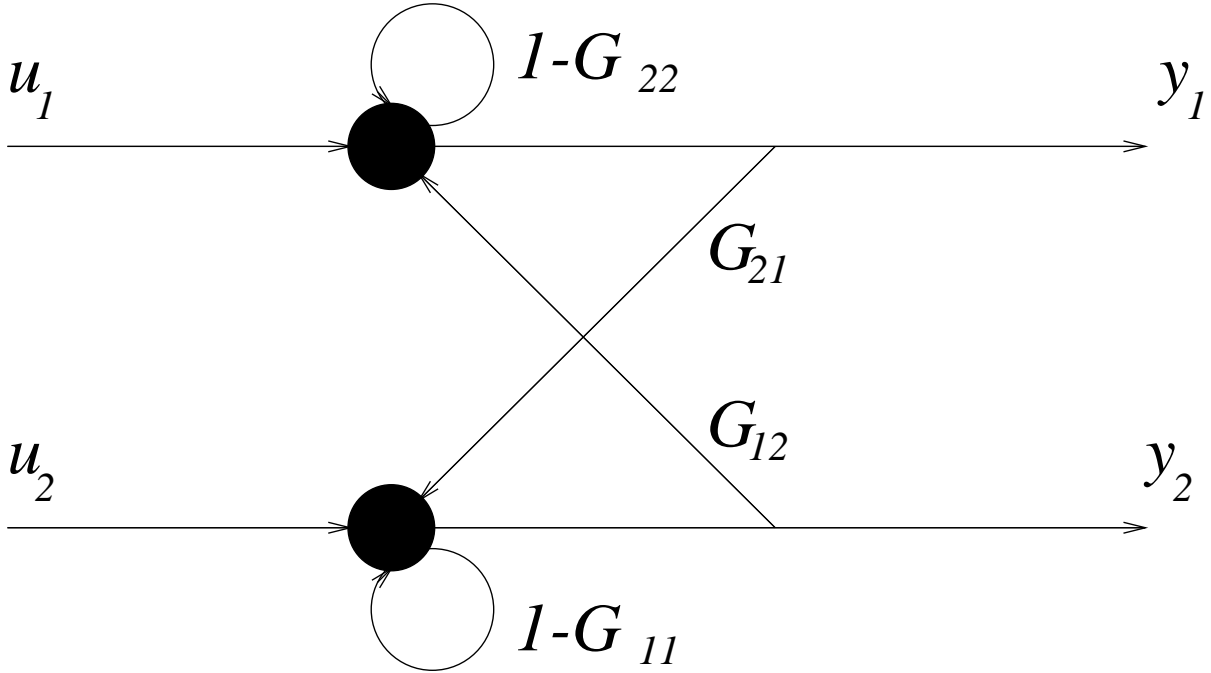,width=16 true cm}}
\caption{Two-neuron normalized network with lateral inhibition.}
\end{figure}
In addition to the lateral inhibition weights $G_{ij}$, $i\neq
j$, self-connections are also included (with their weights
defined so as to conform with our definition of $\gss{G}{}{}$ as
the separating matrix). The latter allow the autocorrelations to
be normalized. Thus, if $\gss{y}{}{}=\bmx{cc} y_{1} &
y_{2}\emx^{T}$ is to be spatially white, with
$E[\gss{y}{}{}\gss{y}{}{T}]=\gss{I}{}{},$ the following,
normalized version of the rule \pref{eq:lHebb} is preferable
\cite{g99}: \beq \Delta \gss{G}{}{}\propto
\gss{I}{}{}-\gss{y}{}{}\gss{y}{}{T}. \label{eq:lnHebb} \eeq
However, in order to do more than decorrelating the components of
$\gss{y}{}{}$, one needs to resort to HOS, which is implicitly
achieved by inserting nonlinear functions in the above rule. The
result is \beq \Delta\gss{G}{}{}\propto
\gss{I}{}{}-\ga{f}{\gss{y}{}{}}\ga{g}{\gss{y}{}{T}}
\label{eq:nlHebb}\eeq which implements the (normalized) nonlinear
anti-Hebbian learning \cite{g99}. Similar algorithms, in their
equivariant form: \beq \Delta\gss{G}{}{}\propto
\left(\gss{I}{}{}-\ga{f}{\gss{y}{}{}}\ga{g}{\gss{y}{}{T}}\right)\gss{G}{}{},
\label{eq:fg} \eeq have been derived in the BSS context
\cite{cur94}. Note that \pref{eq:equivInfomax} results from
\pref{eq:fg} with $\ga{g}{\gss{y}{}{}}=\gss{y}{}{}.$ Lateral
inhibition has been also used in \cite{fg99} for blindly
deconvolving MIMO systems. The network of Fig.~3, with the
self-connection weights set to zero, and the adjusting rule
\[
\Delta \gss{G}{}{}\propto -\ga{f}{\gss{y}{}{}}\ga{g}{\gss{y}{}{T}}
\]
yields the Jutten-H\'{e}rault neuromimetic architecture, one of
the earliest BSS approaches \cite{jh91,g99}.

\subsection{Stability}
It follows from \pref{eq:equivInfomax} that the stationary points
of $J$ will satisfy: \beq
E[\ga{f}{\gss{y}{}{}}\gss{y}{}{T}]=\gss{I}{}{}. \label{eq:statp}
\eeq Clearly, \pref{eq:statp} is satisfied by the desired
solution $\gss{y}{}{}=\gss{a}{}{}$ (i.e.,
$\gss{G}{}{}=\gss{H}{}{-1}$) since then
$E[f_{i}(y_{i})y_{j}]=E[f_{i}(y_{i})]E[y_{j}]=0$ for $i\neq j$ in
view of the independence of the sources \cite{p84}. Moreover, we
may assume a normalization such that $E[f_{i}(y_{i})y_{i}]=1$ or
alternatively replace $\gss{I}{}{}$ above by a diagonal matrix
with diagonal elements $E[f_{i}(a_{i})a_{i}]$ \cite{ac98}.
Eq.~\pref{eq:statp} would then be written in the form: \beq
E[\ga{f}{\gss{y}{}{}}\gss{y}{}{T}]=E[\gss{y}{}{}\gss{y}{}{T}]
\label{eq:f(y)y=yy}\eeq known as the {\em Bussgang condition}
\cite{b52,b94}.\footnote{The close connection of the Bussgang
algorithms with the IT approaches discussed here has been
recognized in \cite{lgbs00}. See also \cite{m00} for an
interesting link with IT criteria for blind channel estimation in
digital communications.} Notice that a linear choice for
$\ga{f}{\cdot}$, for example $\ga{f}{\gss{y}{}{}}=\gss{y}{}{}$,
would fail to separate the sources, since, as one can see from
\pref{eq:statp}, only independence to second order
(decorrelation) would then be achievable. It is of interest to
note that if the $\phi_{i}$'s are pdf's of Gaussian sources, the
corresponding score functions will be linear.

Although the desired solutions are among the fixed-points of the
recursion \pref{eq:equivInfomax}, it will not converge to them
unless they are stable. Necessary and sufficient stability
conditions were derived in \cite{acc97} and demonstrate the fact
that the separation performance depends on the source statistics
in conjunction with the choice of the nonlinearities employed. Let
\beqs \sigma_{i}^{2} & = & E[y_{i}^{2}] \\
k_{i} & = & E[f_{i}^{\prime}(y_{i})] \\
m_{i} & = & E[y_{i}^{2}f_{i}^{\prime}(y_{i})]. \eeqs Then the
stability conditions may be stated as \cite{acc97}: \beqa m_{i}+1
& > & 0 \\
k_{i} & > & 0 \\
\sigma_{i}^{2}\sigma_{j}^{2}k_{i}k_{j} & > & 1 \mbox{\ for all\
}i\neq j. \eeqa One can see that the above conditions imply the
following \beq (1+\kappa_{i})(1+\kappa_{j})>1,
\label{eq:kappas}\eeq where \beq
\kappa_{i}=E[f_{i}^{\prime}(y_{i})]E[y_{i}^{2}]-E[f_{i}(y_{i})y_{i}],
\eeq derived independently in \cite{cl96}. Since
$E[f_{i}^{\prime}(a_{i})]=\frac{E[f_{i}(a_{i})a_{i}]}{E[a_{i}^{2}]}$
for Gaussian $a_{i}$, $\kappa_{i}$ turns out to be zero in that
case. Hence if two sources, say $a_{i}$ and $a_{j}$ are Gaussian,
then $\kappa_{i}=\kappa_{j}=0$ and the condition \pref{eq:kappas}
is not satisfied. This simple argument leads us once more to the
constraint of at most one Gaussian source.

The above conditions provide us with a quantitative measure of
the permitted mismatch between the functions $\phi_{i}$ in
\pref{eq:uniBSS} and the source pdf's \cite{car98a}. Namely,
$f_{i}$'s need to be so chosen as to satisfy \pref{eq:kappas}. An
important categorization of the source statistics is with respect
to their relation to the normal ones. A pdf which is relatively
flat (e.g., uniform) is called {\em sub-Gaussian.} If it has a
spiky appearance with heavy tails (e.g., Laplace), it is said to
be {\em super-Gaussian.} A parameterization of these pdf types as
members of the family of exponential densities is given in
\cite{bgr80}. Fig.~4 shows an example of a sub-Gaussian and a
super-Gaussian pdf.
\begin{figure}
\centerline{\psfig{figure=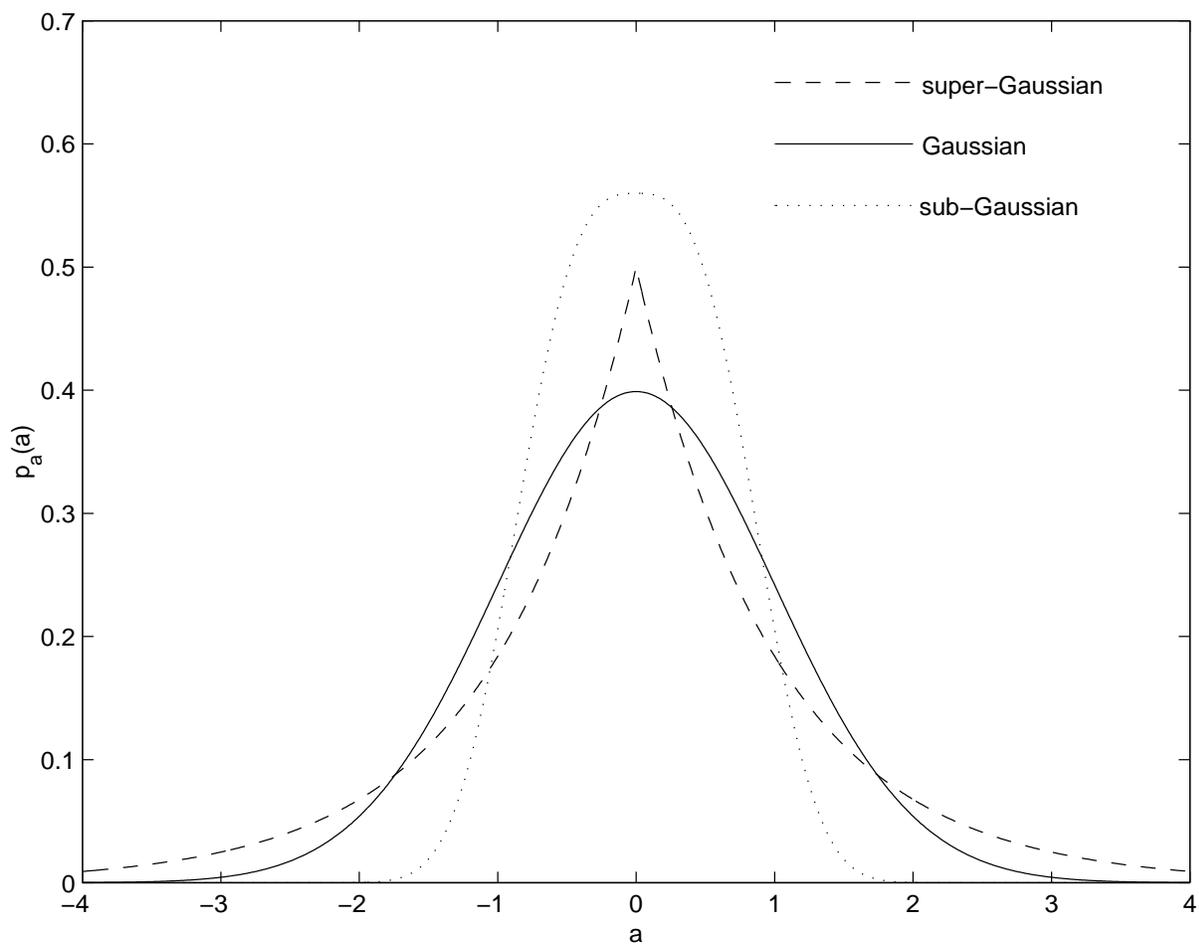,width=16 true cm}}
\caption{Examples of a sub-Gaussian, a super-Gaussian, and a
Gaussian pdf.}
\end{figure}
Whether a pdf is sub-Gaussian, Gaussian, or super-Gaussian, is
commonly seen in the sign of its 4th-order cumulant, $c_{4}.$ It
will be respectively, negative, zero, and positive. Data in
digital communications are usually sub-Gaussian. Examples of
super-Gaussian sources are found in speech and image, among
others. The Infomax algorithm of \cite{bs95} using the logistic
nonlinearity was seen to be successful in separating
super-Gaussian sources. Instead, it failed for sub-Gaussian
mixtures. This is because the nonlinearity $\ga{g}{\cdot}$ that
was used did not match sufficiently well the sub-Gaussian cdf so
that the requirements for stability be met. A simple choice for
the score functions corresponding to sub-Gaussian sources is
based on the cubic nonlinearity, e.g., $f_{i}(y_{i})=y_{i}^{3}.$
With this choice, $-\kappa_{i}$ is reduced to $c_{4}(y_{i})$. The
stability follows since $-\kappa_{i}<0$ is sufficient for
\pref{eq:kappas}.

However, when no information on the source statistics is
available or the sources are mixed sub- and super-Gaussian, such
simple choices are inadequate. Even if one knows which sources
are sub- and which are super-Gaussian, there is the problem of
unavoidable permutation in the separation result, hence one
cannot be sure of which source a function will correspond to. A
common approach to overcoming this problem is to parameterize the
nonlinearities so as to encompass the expected source
distributions. The relevant parameters are then adapted at the
same time with the separating matrix
\cite{cscos97,ckkv99,ac98,g98,lgs99}. Such could be simply the
sign of the estimated $c_{4}(y_{i})$ \cite{g98} or, in more
sophisticated schemes, a parameter connected with the satisfaction
of the stability conditions \cite{ho98,lgs99}. A related approach
is that of approximating the unknown pdf via a truncated
expansion, yielding a polynomial form whose coefficients are
determined by the source cumulants \cite{car98a}. In
Section~5.3.1 we discussed a similar approach to the
simplification of IT criteria.

Ways of stabilizing an algorithm, that weaken the constraints on
the choice of $\ga{f}{\cdot}$, have also been developed. The
multiplication by the inverse of the Hessian matrix (which is
diagonal when using the natural gradient) is shown in \cite{acc97}
to result in a stabilized version. In \cite{hlk00}, it is shown
that a simple substitution of the term
$\ga{f}{\ga{y}{k}}\gssa{y}{}{T}{k}$ in \pref{eq:equivInfomax} by
its transpose, $\ga{y}{k}\ga{f}{\gssa{y}{}{T}{k}}$, suffices to
correct an instability in the case that all $f_{i}$ coincide.

\section{FastICA and Its Variants}
As already mentioned, the source estimates in algorithms based on
polynomial nonlinearities $f_{i}$ are not robust to outliers. The
criterion \pref{eq:cum4}, involving 4th-order standardized
cumulants is an example of this \cite{h99a}. To address this
problem, the following generalized criterion was proposed
\cite{ho98}: \beq
\max_{G^{T}G=I}\left(\psi_{F}(\gss{G}{}{})\sr{\triangle}{=}
\|E[\ga{F}{\gss{y}{}{}}-\ga{F}{\gss{y}{}{G}}]\|^{2}\right)
\label{eq:Hyv} \eeq where $\gss{y}{}{}=\gss{G}{}{}\gss{u}{}{}$ as
before, $\ga{F}{\cdot}$ is a smooth, nonlinear, non-quadratic
vector function with equal components, $F$, and data prewhitening
is assumed, i.e., $\gss{G}{}{}$ is orthogonal. Recall that
$\gss{y}{}{G}$ denotes a Gaussian random vector with the same
mean and covariance as $\gss{y}{}{}.$ In this case,
$\gss{y}{}{G}$ is standardized, i.e.,
$E[\gss{y}{}{G}(\gss{y}{}{G})^{T}]=\gss{I}{}{}.$ Note that, as in
IT criteria discussed above, the underlying principle here is
again that of maximizing the distance of the output $\gss{y}{}{}$
from a Gaussian signal, using the nonlinear function
$\ga{F}{\cdot}$ to extract its HOS.

Let $\gss{g}{i}{T}$ denote the $i$th row of $\gss{G}{}{}.$ Then
\pref{eq:Hyv} may be equivalently written as: \beqa
\max_{g_{1},\ldots,g_{N}} & &
\sum_{i=1}^{N}\psi_{F}(\gss{g}{i}{}) \nonumber \\
\mbox{subject to\ } & & \gss{g}{i}{T}\gss{g}{j}{}=\delta_{ij}
\label{eq:SHyv} \eeqa where \beq
\psi_{F}(\gss{g}{i}{})\sr{\triangle}{=}
\left\{E[F(\gss{g}{i}{T}\gss{u}{}{})-F(y^{G}_{i})]\right\}^{2}
\label{eq:Hyv1} \eeq and $\delta_{ij}$ is the Kronecker delta.
Since the second term in \pref{eq:Hyv1} is constant, it is at the
extrema of $E[F(y_{i})]$ that correspond the maxima of
$\psi_{F}(\gss{g}{i}{}).$ Whether it will be to a maximum or a
minimum of $E[F(y_{i})$ depends, respectively, on whether
$E[F(y_{i})$ is greater or smaller than $E[F(y_{i}^{G})]$. For
$F(y)=y^{4}$ this translates into $y_{i}$ being super- or
sub-Gaussian, respectively. For a general $F$, these relations
imply generalized definitions of super- and sub-Gaussianity
\cite{rk01}.

It is easier to present the optimization of the above criterion
by focusing on the extraction of a single source. Once a
$\gss{g}{i}{}$ has been determined, the rest of them can be
computed with the additional constraint that they be orthogonal
to that already found. This can be repeated for another
$\gss{g}{j}{}$, and so on. Through this so-called {\em deflation}
approach \cite{dl95}, a single row of $\gss{G}{}{}$ is computed at
each step, reducing the problem size by one. Popular methods for
enforcing the constraint \pref{eq:SHyv} all along this process
include Gram-Schmidt-like orthogonalization schemes
\cite{h99a,p00}. Extracting the sources one by one may be useful
not only in view of the simplification this implies for the
algorithm, but also in applications that ask for the extraction
of only a few of the sources and in a desired order \cite{cc98}.
Furthermore, such an approach may be the only possible solution
in cases where not all sources can be recovered at once, for
example when $\gss{H}{}{}$ is not invertible (see Section~9).

Consider then the problem of maximizing one of the terms in the
sum \pref{eq:SHyv} subject to the constraint that the separating
vector $\gss{g}{}{}$ be of unit norm.\footnote{This is not to be
confused with the vector function $\ga{g}{\cdot}$ defined in
Section~5.} Assume moreover that $E[F(y)]$ is to be maximized.
Then the optimization problem reduces to \beq
\max_{\|g\|=1}E[F(\gss{g}{}{T}\gss{u}{}{})] \label{eq:E(F(y))}\eeq
with the following characterization of its stationary points
\cite{h99a}: \beq
E[\gss{u}{}{}f(\gss{g}{}{T}\gss{u}{}{})]-\beta\gss{g}{}{}=\gss{0}{}{}.
\label{eq:fixedpoint} \eeq $f$ is the derivative of $F$ and the
Lagrange multiplier $\beta$ is given by $\beta=E[yf(y)]$. A
steepest ascent scheme yields the update equation: \beq
\ga{g}{k+1}=\ga{g}{k}+
\mu(k)\left\{E[\gss{u}{}{}f(\gssa{g}{}{T}{k}\gss{u}{}{}]-\beta(k)\ga{g}{k}\right\}.
\label{eq:Hyvgrad} \eeq Via a Newton approximation of the above,
the following algorithm results \cite{h99a}: \beqa
\gssa{g}{}{+}{k+1} & = &
E[\gss{u}{}{}f(\gssa{g}{}{T}{k}\gss{u}{}{})]
-E[f^{\prime}(\gssa{g}{}{T}{k}\gss{u}{}{})]\ga{g}{k}
\label{eq:FastICA1}
\\
\ga{g}{k+1} & = &
\frac{\gssa{g}{}{+}{k+1}}{\|\gssa{g}{}{+}{k+1}\|}
\label{eq:FastICA2}\eeqa known as {\em FastICA}. The normalization
in \pref{eq:FastICA2} is needed to enforce the unit-norm
constraint and stabilize the algorithm. The connection of the
FastICA algorithm, in its multi-source version, with the
algorithm \pref{eq:equivInfomax} was studied in \cite{h99d}. It
was shown that it results as a batch version of
\pref{eq:equivInfomax} via a generalization consisting of the use
of not necessarily the same step size for all rows of
$\gss{G}{}{}.$

Note that the above algorithm may be viewed as resulting from
\pref{eq:Hyvgrad} by choosing the step size as \beq
\mu(k)=\frac{1}{\beta(k)-E[f^{\prime}(\gssa{g}{}{T}{k}\gss{u}{}{})]}.
\label{eq:Hyvmu} \eeq It is claimed in \cite{h99a} that this
choice yields the fastest convergence. In fact, as recently
shown, $\mu$ should instead be chosen as \cite{rk01}: \beq
\mu(k)=\frac{1}{\beta(k)} \label{eq:RKmu} \eeq resulting in the
following, faster variation of
\pref{eq:FastICA1}--\pref{eq:FastICA2}: \beqa \gssa{g}{}{+}{k+1}
& = & E[\gss{u}{}{}f(\gssa{g}{}{T}{k}\gss{u}{}{})] \label{eq:RK1}
\\
\ga{g}{k+1} & = &
\frac{\gssa{g}{}{+}{k+1}}{\|\gssa{g}{}{+}{k+1}\|}. \label{eq:RK2}
 \eeqa

The latter algorithm can be viewed as a solution approach to
eq.~\pref{eq:fixedpoint} via a fixed-point iteration. In fact,
for $F(y)=y^{2p}$ and hence $f(y)=2py^{2p-1}$, it is readily seen
to reduce to the so-called {\em super-exponential algorithm
(SEA)}, well-known from the blind deconvolution problem
\cite{sw93,sw94}. An interpretation of the SEA as a gradient
optimization scheme can be found in \cite{mr00}. It is the
fastest known method for maximizing the ratio
$\left|\frac{c_{2p}(y)}{(c_{2}(y))^{p}}\right|$, a criterion known
as the {\em minimum-entropy} or {\em Donoho's} criterion
\cite{d81,w90,c96,rm99a}. In practice $p$ is usually set to two.
In that case, $c_{2p}(y)=c_{4}(y)=E[y^{4}]-3(E[y^{2}])^{2}$, and
for standardized $y$, $c_{4}(y)=E[y^{4}]-3.$ The connection with
\pref{eq:E(F(y))} when $F(y)=y^{4}$ is thus evident. Note also
that for this choice of $F$ the criterion \pref{eq:SHyv} reduces
to \pref{eq:cum4}.

Donoho's criterion has been shown to be equivalent to the
so-called {\em constant modulus (CM)} criterion, which for $p=2$
reads \cite{j98,j00} \beq \min_{g}E[(y^{2}-1)^{2}]. \label{eq:CM}
\eeq The equivalence holds for $\gss{g}{}{}$ of optimized norm in
\pref{eq:CM} \cite{r99}. Applying a stochastic gradient descent
to this criterion results in the celebrated {\em Constant Modulus
Algorithm (CMA)} \cite{g80,ta83}: \beq \ga{g}{k+1}=\ga{g}{k}-\mu
(y^{2}(k)-1)y(k)\ga{u}{k} \label{eq:CMA}\eeq which has been and
probably will continue to be one of the major subjects of the
blind equalization research \cite{j98,j00}. It can be derived as a
special case of the algorithm \pref{eq:equivInfomax} with
$f(y)=y^{3}$ and hence, as we saw in Section~6.3, it can only be
applied to sub-Gaussian sources. Modified versions that can
process super-Gaussian sources can be derived with the aid of the
equivalence mentioned above, in view of the ability of the SEA to
cope with both kinds of inputs \cite{rm99b}.

\section{Algebraic Approaches}
So far we have been concerned with BSS approaches that stem from
optimizing a criterion via a gradient recursive scheme. A great
number of methods follow instead an algebraic approach to solving
the problem. Although, as we will see shortly, they can also be
interpreted as criteria-based schemes, it is their point of view
and the tools they employ that differentiate them from the
methods discussed above.

\subsection{Statistical Approaches}
What underlies the algebraic schemes employing statistical
quantities is the exploitation of the fact that independence of
the components of a vector random process is equivalent to all its
{\em cross-cumulants} being null. This implies that BSS can be
achieved by setting the cross-cumulants of $\gss{y}{}{}$ equal to
zero and solving the resulting equations for $\gss{G}{}{}.$ Such
methods have been reported in e.g. \cite{lac97,yi00}. They get too
complicated however for realistic problem sizes. Therefore, more
efficient and less direct methods are required.

The basic quantity in stochastic algebraic approaches is the {\em
cumulant tensor} \cite{m87}. By {\em tensor} we mean here a
$q$-way array, i.e., one whose elements are addressed via $q$
indices \cite{b95,c00b}. It is the 4th-order cumulant of the
$N$-vector $\gss{y}{}{}$ that is most commonly
encountered.\footnote{Third-order cumulants may not be applicable
since they null out when signals with symmetric distributions are
involved \cite{c94a}.} It constitutes a 4th-order tensor,
$\gss{\cal C}{4,y}{}$, with dimensions $N\times N\times N\times
N$ \cite{m87}. The well-known symmetry property of cumulants
\cite{lac97,m91} implies that $({\cal C}_{4,y})_{i,j,k,l}$ does
not depend on the order of its indices. Such a tensor is termed
{\em super-symmetric} \cite{kr01b}.

Another property of cumulants that is employed is that of {\em
multilinearity} \cite{m91,nm93}. This refers to the way the
cumulants of the output of a linear system are related to those
of its input. For the mixing system \pref{eq:u=Ha} it takes the
form: \beq {\rm
cum}(u_{i},u_{j},u_{k},u_{l})=\sum_{i_{1},i_{2},i_{3},i_{4}}{\rm
cum}(a_{i_{1}},a_{i_{2}},a_{i_{3}},a_{i_{4}})H_{i,i_{1}}H_{j,i_{2}}H_{k,i_{3}}H_{l,i_{4}}.
\label{eq:multil} \eeq Using the Tucker product notation
\cite{gr99}, this may be written in the more compact way: \beq
\gss{\cal C}{4,u}{}=\gss{H}{}{}\sr{{\cal
C}_{4,a}}{\star}\gss{H}{}{}\sr{{\cal
C}_{4,a}}{\star}\gss{H}{}{}\sr{{\cal C}_{4,a}}{\star}\gss{H}{}{}
\label{eq:Cu=HHHHCa} \eeq where $\gss{\cal C}{4,a}{}$ is diagonal
because of the independence of the sources. The above
decomposition can thus be viewed as a 4th-order eigenvalue
decomposition (EVD) of the 4-way array $\gss{\cal C}{4,u}{}$
\cite{car90a,car90b} and could be used to blindly identify
$\gss{H}{}{}.$ In contrast to its 2nd-order counterpart,
eq.~\pref{eq:Ru=HH^T}, \pref{eq:Cu=HHHHCa} does not suffer from
nonuniqueness problems in general. This point will be elaborated
upon in the next section.

For the demixing model \pref{eq:y=Gu} one may analogously write:
\beq \gss{\cal C}{4,y}{}=\gss{G}{}{}\sr{{\cal
C}_{4,u}}{\star}\gss{G}{}{}\sr{{\cal
C}_{4,u}}{\star}\gss{G}{}{}\sr{{\cal C}_{4,u}}{\star}\gss{G}{}{}
\label{eq:Cy=GGGGCu} \eeq or, in terms of the global system
$\gss{S}{}{}=\gss{G}{}{}\gss{H}{}{}$ \cite{gr99}: \beq \gss{\cal
C}{4,y}{}=\gss{S}{}{}\sr{{\cal
C}_{4,a}}{\star}\gss{S}{}{}\sr{{\cal
C}_{4,a}}{\star}\gss{S}{}{}\sr{{\cal C}_{4,a}}{\star}\gss{S}{}{}.
\label{eq:Gy=SSSSCa} \eeq Assume $\gss{u}{}{}$ has been
standardized. It can be shown that an orthogonal $\gss{G}{}{}$ in
\pref{eq:Cy=GGGGCu} leaves the norm of $\gss{\cal C}{4,u}{}$
unchanged\footnote{The Frobenius norm is meant here \cite{l97}.}:
\beq \|\gss{\cal C}{4,y}{}\|=\|\gss{\cal C}{4,u}{}\|.
\label{eq:|Cy|=|Cu|} \eeq This has very important implications
for contrasts of the type given in \pref{eq:cum4}. It follows
that maximizing the sum of the autocumulants squared is
equivalent to minimizing the sum of the cross-cumulants squared.
This brings us back to the cross-cumulant nulling idea discussed
above and shows \pref{eq:cum4} to be a cumulant tensor
diagonalization criterion \cite{c94b,cc96}.

It has been shown that for $N=2$ sources the above maximization
can be given a closed-form solution, since $\gss{G}{}{}$ is then
a simple Givens rotation matrix \cite{c94a}. For $N>2$ a
Jacobi-type algorithm \cite{gv96} was proposed that iteratively
processes the components of $\gss{y}{}{}$ in pairs \cite{c94a}.
In addition to being simple and efficient, this method relies on
the theoretical result that pairwise independence, under some weak
constraints, is sufficient for ICA \cite{c94a,caoliu}. The
resulting algorithm has been seen to be a batch version of the
nonlinear PCA rule presented in Section~6 \cite{g99}. Though
proven to be effective in practice, no theoretical proof of its
global convergence has been devised yet \cite{c93}. The same
basic idea, of employing iterative Jacobi rotations to effect a
diagonalization, can be found in a number of other approaches as
well, e.g., \cite{cs93,vp96,s98}.

A method for indirectly diagonalizing $\gss{\cal C}{4,u}{}$, via
a {\em joint} diagonalization of a set of related matrices was
proposed in \cite{cs93}. It can in fact be shown that a single
matrix suffices, provided it is properly chosen \cite{tlsh91}.
However, since this choice is not easy to make due to the blind
nature of the problem, a set of matrices is normally employed. An
example could be the set of slices of $\gss{\cal C}{4,u}{}$
\cite{cs93}. It turns out however that the eigematrices of
$\gss{\cal C}{4,u}{}$, resulting from the EVD of the
corresponding symmetric operator, provide a much smaller
sufficient set. This leads to an efficient algorithm, known as
{\em Joint Approximate Diagonalization of Eigenmatrices (JADE)}
\cite{cs93,v98}. It can be seen to correspond to the optimization
of a cost similar to \pref{eq:cum4}, in which only the
cross-cumulants having their first and second indices different
are minimized. An even weaker criterion, that proves nevertheless
to be effective, was proposed in \cite{lmv00d}.

Another way of obtaining the decomposition \pref{eq:Cu=HHHHCa} is
via an extension of the EVD to tensors, called {\em Higher-Order
EigenValue Decomposition (HO-EVD)} and proposed and studied in
\cite{lmv94,l97,lmv00b}. It {\em approximates} \pref{eq:Cu=HHHHCa}
with the aid of a singular value decomposition (SVD) of $\gss{\cal
C}{4,u}{}$ unfolded to a $N\times N^{3}$ matrix. The so-called
{\em core} tensor however, approximating $\gss{\cal C}{4,a}{}$,
is not guaranteed to be diagonal, unless of course the given data
exactly satisfy the assumed model and the cumulant estimates are
exact, which is rarely the case. Hence, this method usually serves
to provide a good initialization to some other, iterative
algorithm. See \cite{lmv00d} for an interpretation of HO-EVD as a
cost-optimizing method.

\subsection{Deterministic Approaches}
The methods described above work with statistical quantities,
estimated from the given data. Approaches that act directly on
the data also exist and could be referred to as {\em
deterministic.} The best way of conveying the basic idea is by
looking at algebraic methods for solving the CM cost minimization
problem. Note that the Kronecker product allows us to write the
square of $y=\gss{g}{}{T}\gss{u}{}{}$ in the form \beq
y^{2}=(\gss{g}{}{}\otimes\gss{g}{}{})^{T}(\gss{u}{}{}\otimes\gss{u}{}{}).
\label{eq:y2=gguu} \eeq Let us take a block of observation
vectors, say $\ga{u}{1},\ga{u}{2},\ldots,\ga{u}{T}$. Observe that
the minimum CM criterion \pref{eq:CM} aims at approximating
$y^{2}$ by one. In view of \pref{eq:y2=gguu} this can be stated
as the algebraic problem of decomposing the solution,
$\gss{w}{}{}$, to the following $T\times N^{2}$ linear system of
equations \beq
\underbrace{\bmx{c} (\ga{u}{1}\otimes\ga{u}{1})^{T} \\
(\ga{u}{2}\otimes\ga{u}{2})^{T}
\\ \vdots \\ (\ga{u}{T}\otimes\ga{u}{T})^{T} \emx}_{\gss{P}{}{}}
\gss{w}{}{}=\underbrace{\bmx{c} 1 \\ 1 \\ \vdots \\ 1
\emx}_{\gss{1}{}{}} \label{eq:Pw=1} \eeq in the Kronecker square
form $\gss{w}{}{}=\gss{g}{}{}\otimes\gss{g}{}{}$
\cite{vp96,gc99,gc00}.

Note that \pref{eq:Pw=1} is not square and for $T>N^{2}$ (common
assumption) is overdetermined. A solution $\gss{w}{}{}$ is not in
general decomposable as above. \cite{gc99,gc00} consider solving
\pref{eq:Pw=1} in the LS sense and then approximating the result
by a Kronecker square to yield the separation filter. A more
sophisticated approach is followed by the {\em Analytical
Constant Modulus Algorithm (ACMA)} \cite{vp96,v01} and applied to
the multi-source extraction problem. A basis,
$\{\gss{w}{1}{},\gss{w}{2}{},\ldots,\gss{w}{N}{}\}$, for the
solution set of \pref{eq:Pw=1} is first computed and then
approximated by another whose vectors have the form
$\gss{g}{}{}\otimes\gss{g}{}{}.$ This approximation is seen to
lead again to a joint diagonalization problem, involving the
matrices $\gss{W}{i}{}={\rm unvec}(\gss{w}{i}{}).$

Deterministic approaches have the advantage of requiring only a
few data to provide a good estimate for $\gss{G}{}{}$
\cite{vp96}. Interestingly, the ACMA solution was shown to
asymptotically converge to the CM one, when $T\rightarrow\infty$
\cite{v01}.

\section{Undermodeling: Problems and Solutions}
We have thus far assumed that $\gss{H}{}{}$ is square, or as it
is otherwise said, that there as many sensors as sources. A more
realistic scenario would involve a rectangular mixing matrix, of
dimensions $M\times N$ with $M$ not necessarily equal to $N$. In
this section we will be concerned with the problems that emerge
in such a case and possible approaches to their solution.

First, let us consider the case of more sensors than sources,
with a full column rank mixing matrix. This is the case studied in
the majority of the BSS works as it presents no particular
difficulties compared to that of square $\gss{H}{}{}.$ The reason
for this, as already explained in Section~2, is that an
$\gss{H}{}{}$ with more rows than columns can be easily inverted,
provided of course that it be of full column rank. Moreover, this
case reduces to the simpler one as a result of a sphereing
operation. Indeed, if $\gss{H}{}{}$ is $M\times N$ with $M>N$ in
\pref{eq:Ru=HH^T}, an EVD of $\gss{R}{u}{}$ will yield an
orthonormal basis for its range, represented by a columnwise
orthogonal matrix $\gss{A}{}{}$, revealing at the same time the
number of sources, $N$. Therefore, there will be an $N\times N$
orthogonal matrix $\gss{Q}{}{T}$ such that
$\gss{A}{}{}=\gss{H}{}{}\gss{Q}{}{T}.$ Transforming the mixture
data as $\ga{\bar{u}}{n}=\gss{A}{}{T}\ga{u}{n}$ will then result
in the $N$-source, $N$-sensor system
$\ga{\bar{u}}{n}=\gss{Q}{}{}\ga{a}{n}$ with an orthogonal mixing
matrix. From this point on, a method for square systems can be
applied.

The case of less sensors than sources is however much more
difficult to be studied. Recall from Section~2 that this includes
the existence of noise sources as a special case. A consequence
of the noninvertibility of $\gss{H}{}{}$ is that, in contrast to
the case of $M\geq N$, the mixing system identification and
source extraction are two different problems. Even if
$\gss{H}{}{}$ is perfectly known, it is not clear how it could be
inverted to recover $\gss{a}{}{}.$\footnote{This holds for the
general case where no additional structure is known to exist
within $\gss{H}{}{}.$ For example, more sources than sensors are
shown to be relatively easily extracted from a decomposition of
$\gss{\cal C}{4,u}{}$ for the case of a uniform sensor array of
known geometry \cite{sg93}.} It was recently proved \cite{tj99a}
that $\gss{H}{}{}$ is uniquely determined if there are no
Gaussian sources.\footnote{Similar results can be found in
\cite{caoliu}.} Furthermore, the source vector can be extracted
only up to an unknown additive noise vector \cite{tj99a}.

\subsection{Identification}
If $M<N$, $\gss{H}{}{}$ cannot have linearly independent columns.
It can however be that the projectors on these columns be
linearly independent. This means that the $M^{2}\times N$ matrix
$\gss{H}{}{}\odot\gss{H}{}{}$, where $\odot$ denotes the
Kronecker columnwise product \cite{sb01}, can be of full column
rank if $M^{2}>N.$ Based on this assumption, \cite{car91} develops
an algebraic algorithm for identifying $\gss{H}{}{}.$ The
decomposition of $\gss{\cal C}{4,u}{}$ in its eigenmatrices
mentioned above is a basic step in this algorithm. In fact, the
method of \cite{car91} reduces to that for the case of $M\geq N.$
The maximum number of sources the algorithm can afford to can be
shown to be of the order of $M^{2}$ \cite{car91}.

The result is the decomposition of the tensor $\gss{\cal
C}{4,u}{}$ in a sum of $N$ linearly independent rank-1 tensors,
that is 4th-order outer products, as in: \beq \gss{\cal
C}{4,u}{}=c_{4}(a_{1})\gss{h}{1}{}\star\gss{h}{1}{}\star\gss{h}{1}{}\star\gss{h}{1}{}+
c_{4}(a_{2})\gss{h}{2}{}\star\gss{h}{2}{}\star\gss{h}{2}{}\star\gss{h}{2}{}+\cdots
+
c_{4}(a_{N})\gss{h}{N}{}\star\gss{h}{N}{}\star\gss{h}{N}{}\star\gss{h}{N}{}
\label{eq:sPARAFAC} \eeq where $\gss{h}{j}{}$ is the $j$th column
of $\gss{H}{}{}$ and
\[
\left(\gss{h}{j}{}\star\gss{h}{j}{}\star\gss{h}{j}{}\star
\gss{h}{j}{}\right)_{i_{1},i_{2},i_{3},i_{4}}\sr{\triangle}{=}
H_{i_{1},j}H_{i_{2},j}H_{i_{3},j}H_{i_{4},j}.
\]
We say that this tensor has {\em rank} $N$ since it cannot be
decomposed in less than $N$ such terms. Eq.~\pref{eq:sPARAFAC} is
an instance of a decomposition model for general, not necessarily
super-symmetric tensors, known in multi-way data analysis with the
name {\em Parallel Factor Analysis (PARAFAC)} \cite{b97} (or {\em
CANonical DECOMPosition (CANDECOMP)} \cite{l97}). For a 4th-order
$M_{1}\times M_{2}\times M_{3}\times M_{4}$ tensor $\gss{\cal
A}{}{}$ it takes the form: \beq \gss{\cal
A}{}{}=\gss{U}{}{(1)}\sr{\cal S}{\star}\gss{U}{}{(2)}\sr{\cal
S}{\star}\gss{U}{}{(3)}\sr{\cal S}{\star}\gss{U}{}{(4)}
\label{eq:PARAFAC} \eeq where $\gss{U}{}{(i)}$ is an $M_{i}\times
r$ matrix, $r={\rm rank}(\gss{\cal A}{}{})$ and $\gss{\cal
S}{}{}$ is an $r\times r\times r\times r$ diagonal tensor. The
PARAFAC of a 3rd-order tensor is schematically shown in Fig.~5.
\begin{figure}
\centerline{\psfig{figure=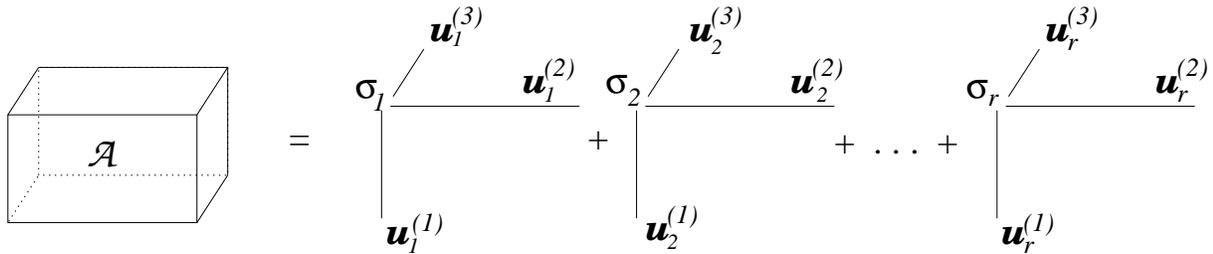,width=16 true cm}}
\caption{PARAFAC analysis of a 3rd-order tensor.}
\end{figure}
PARAFAC extends to higher-order tensors the well-known expansion
of a matrix $\gss{A}{}{}$ in a sum of ${\rm rank}(\gss{A}{}{})$
rank-1 terms \cite{gv96}. In contrast to what holds in
second-order tensors however, this decomposition enjoys the
property of uniqueness under some rather weak conditions. This
fact, along with the possibility of interpreting the BSS problem
as one of determining the independent factors generating the
observed data \cite{ep98}, led to a recent research activity for
applying it to blind signal processing problems
\cite{sb01}.\footnote{A surprising result from the application of
PARAFAC to blind DS-CDMA identification and equalization was
recently reported. 
Not only the channels and their
inputs can be determined but also the spreading codes themselves!}
A sufficient but not necessary condition for uniqueness of the
factors has been first derived in \cite{k77} for 3rd-order
tensors and recently extended to arrays of higher orders
\cite{sb00}. For the tensor in \pref{eq:sPARAFAC} this condition
leads, generically, to \beq 4M\geq 2N+3. \eeq This means that the
uniqueness will be ensured for roughly up to $2M$ sources. This
should be compared with the $M^{2}$ sources that are possible in
the method of \cite{car91}. Note however that the condition of
\cite{sb00} applies to general tensors, not only to cumulants of
linear mixtures.

For a tensor that exactly satisfies a PARAFAC of rank $N$,
methods based on joint diagonalization of matrices have been
derived; see, e.g., \cite{l97}.\footnote{These procedures turn
out to be analogous to algorithms that are classical in array
signal processing \cite{sb01}.} Nonetheless, exact decomposition
may not be possible in practice and it will have to be
approximated. {\em Alternating Least Squares (ALS)} provide the
most common algorithm for this task, referred to sometimes as the
PARAFAC ALS algorithm \cite{b97,bsg99}. The idea is to
alternately optimize the matrix $\gss{U}{}{(i)}$ in every
dimension, in the LS sense, considering the other matrices fixed.
For a super-symmetric tensor, such as that given in
\pref{eq:PARAFAC}, the same method applies without imposing the
equality of these matrices. The final approximation however turns
out to be symmetric. This procedure is provenly convergent,
though not necessarily to a good solution. For super-symmetric
tensors, HO-EVD yields an initialization which has been observed
to be effective in leading to good approximations
\cite{l97}.\footnote{A tensorial equivalent of SVD, known as
HO-SVD, can be applied in the general, non-symmetric case
\cite{lmv00b}.}

An approach to computing the PARAFAC of a super-symmetric tensor
that respects the symmetry property and relies on the
correspondence between super-symmetric tensors and homogeneous
multivariate polynomials was proposed in \cite{cm96}. It is shown
that, in this new domain, the PARAFAC problem translates into
writing a polynomial as a sum of powers of linear forms
\cite{cm96}. Despite the sound theoretical foundations of this
approach however, working algorithms could only be developed for
small-sized problems \cite{cm96,c98,lmv00a}.

\subsection{Inversion}
As already explained, no {\em linear} inversion is possible in
general for a system with more sources than sensors.
Nevertheless, it was proved in \cite{tj99a} that inversion is
possible in case that the sources take values from a finite set.
\cite{c98} presents an algorithm for {\em nonlinearly} inverting
an identified mixing system with discrete inputs. The basic idea
is that discreteness of signals with alphabets as those used in
digital communications can be expressed in terms of polynomial
equations. Writing appropriate polynomial equations for the
$u_{i}$'s in terms of $\gss{H}{}{}$ and the $a_{i}$'s and using
the equations characterizing the discrete range of the latter, a
nonlinear system of equations in the $a_{i}$'s results.
Hopefully, this will not contain many nonlinear combinations of
the sources. These are considered as additional unknowns so the
system becomes linear and can be solved with a LS method. The
`nonlinear' unknowns are then discarded from the solution.

An alternative way out is to follow a deflation approach,
extracting the sources one by one \cite{caoliu}. Note that this
problem can be viewed as a particular case of the PARAFAC
decomposition of $\gss{\cal C}{4,u}{}$ where only a rank-1 term
is computed at a time. For the tensor $\gss{\cal C}{4,u}{}$ this
yields the Donoho's criterion \cite{kr01a}. The ALS algorithm
described above reduces to a higher-order extension of the
well-known power method for rank-1 matrix approximation
\cite{gv96}, known as {\em higher-order power method (HO-PM)}
\cite{lcmv95,lmv00c}. A version of the HO-PM, adapted to
super-symmetric tensors, was recently introduced
\cite{kr01a,kr01b}. Interestingly, the SEA of \cite{rm99b} turns
out to be nothing but this method applied to the tensor
$\gss{\cal C}{4,u}{}$ \cite{kr01a}.

Donoho's cost functional, \beq
\psi_{D}(\gss{g}{}{})\sr{\triangle}{=}\left|\frac{c_{4}(y)}{[c_{2}(y)]^{2}}\right|,
\mbox{\ \ } y=\gss{g}{}{T}\gss{u}{}{} \label{eq:Donoho}\eeq is
multimodal. Most importantly, its local extrema might correspond
to separating filters of a very low performance compared to the
globally optimal ones. This issue has been extensively studied,
especially in the context of CM equalization \cite{j98,j00}. To
deal with this problem, one would like to have a good initial
estimate for $\gss{g}{}{}$ that would lie in the basin of
attraction of a globally optimal solution. This would lead the
gradient algorithm to a separator of acceptable performance. Such
an initialization scheme was devised in \cite{rk00a} and consists
of two successive matrix rank-1 approximations (data prewhitening
is assumed) :
\begin{enumerate}
\item First build from $\gss{\cal C}{4,u}{}$ the $M^{2}\times
M^{2}$ symmetric matrix $\gss{C}{u}{}$ by combining the first two
dimensions in the row dimension and the other two in the column
dimension.
\item Compute the best in the LS sense rank-1 approximation of
$\gss{C}{u}{}$:
\[
\hgss{C}{u}{}=\lambda\gss{w}{}{}\gss{w}{}{T}.
\]
\item Compute the best in the LS sense rank-1 approximation of the
$M\times M$ symmetric\footnote{A proof of the symmetry of
$\gss{W}{}{}$ is given in \cite{kr01b}.} matrix $\gss{W}{}{}={\rm
unvec}(\gss{w}{}{})$:
\[
\hgss{W}{}{}=\varsigma\gss{g}{}{0}(\gss{g}{}{0})^{T}
\]
\item Initialize the SEA with $\gss{g}{}{0}.$
\end{enumerate}
Note the similarity with the deterministic approaches discussed
in Section~8.2. This method has been seen via extensive
simulations to frequently outperform the HO-EVD initialization
scheme in terms of the proximity of the estimate provided to the
globally optimal solution. Fig.~6 depicts the initial estimates
computed with these two methods and the corresponding trajectories
followed by the SEA, for a 3-sensor, 7-source system with mixed
sub- and super-Gaussian sources.
\begin{figure}
\centerline{\psfig{figure=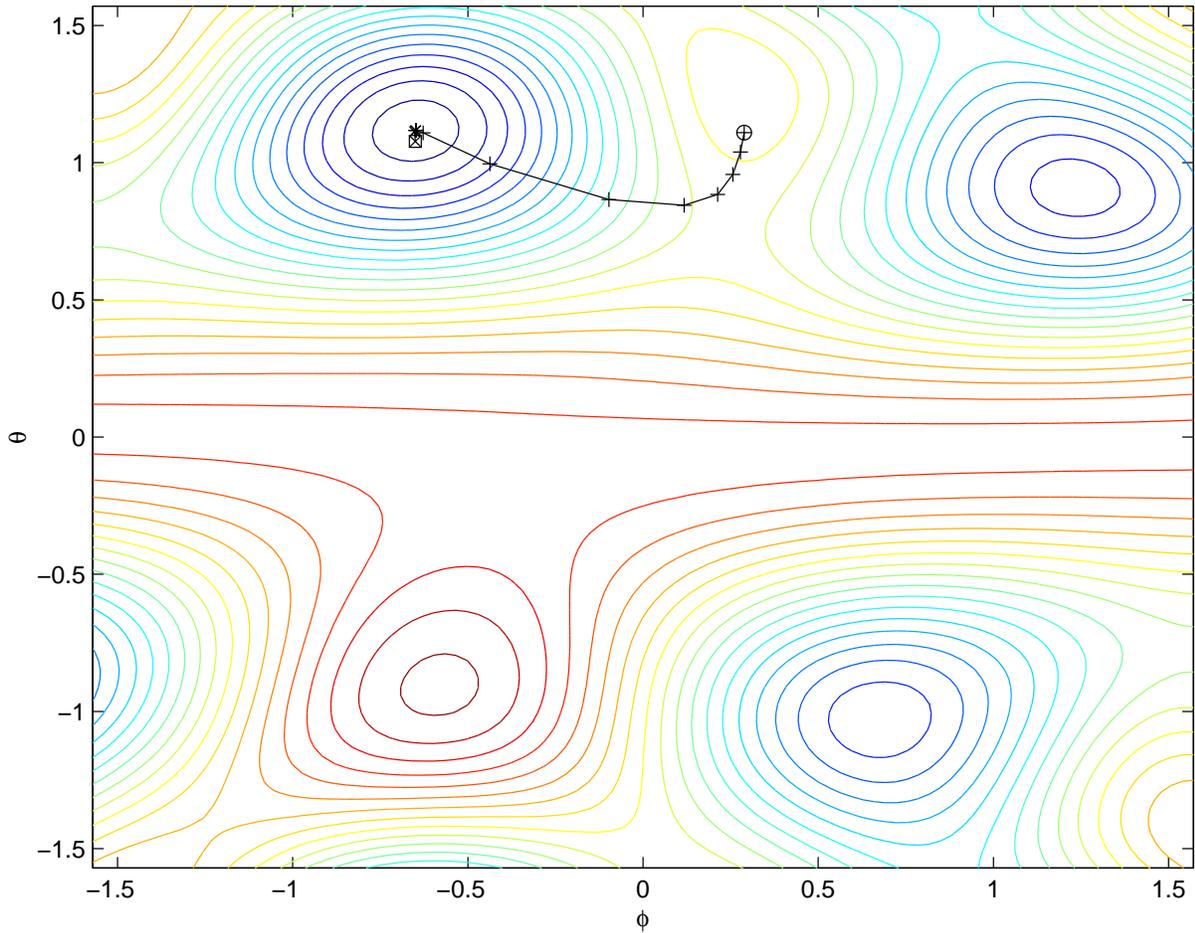,width=16 true cm}}
\caption{Visualization of the SEA for two initialization methods.
The HO-EVD-based estimate is denoted by a small circle and the
trajectory followed by $+$'s. The initial estimate provided by
the method of \cite{rk00a} is denoted by a small square and the
subsequent estimates by $\times$'s. The angles $\theta$ and
$\phi$ parameterize the unit-norm 3-vector $\ga{g}{k}.$}
\end{figure}
As shown in the figure, the initialization $\gss{g}{}{0}$ computed
with the above method lies so close to the global optimum that
SEA iterations are hardly needed. In fact, it can be shown
\cite{rk00b} that $\gss{g}{}{0}$ will equal a perfect separator if
it exists (that is, when $M\geq N$). More generally and for
sources that are all sub- or super-Gaussian, the resulting
separation performance is shown to be bounded as \cite{rk00a}:
\beq \varsigma^{2}|\lambda|\leq \psi_{D}(\gss{g}{}{0})\leq
|\lambda|. \eeq Notice that these bounds are computable {\em
a-priori}, that is, based on only the observation data statistics.

In view of the above discussion, it can be seen that an
approximation to the Donoho's criterion, given as: \beq
\max_{\|g\|=1}\left(\hat{\psi}_{D}(\gss{g}{}{})\sr{\triangle}{=}
\left|\lambda(\gss{g}{}{T}\gss{W}{}{}\gss{g}{}{})^{2}\right|\right),
\label{eq:Donohoapprox} \eeq could be an interesting alternative.
First, this cost is readily verified to be unimodal, in the sense
that all its stable stationary points correspond to separators of
the same quality. Next, it inherits all the good properties of
the initialization given above. An algorithm based on this
criterion would be globally convergent, that is, its convergent
points would be independent of its initialization. Moreover, they
would provide high performance separating filters. Such an
algorithm was developed and successfully tested in \cite{rk00b}
and is described by the following two recursions: \beqa
\ga{W}{k+1} & = &
\ga{W}{k}+\frac{\mu_{1}}{1+\mu_{1}\|\ga{u}{k}\otimes\ga{u}{k}\|^{2}}
\left(1-\gssa{u}{}{T}{k}\ga{W}{k}\ga{u}{k}\right)\ga{u}{k}\gssa{u}{}{T}{k}
\label{eq:Volterra}
\\
\gssa{g}{}{+}{k+1} & = & \ga{g}{k}+\mu_{2}\ga{W}{k+1}\ga{g}{k} \label{eq:equalizer}\\
\ga{g}{k+1} & = &
\frac{\gssa{g}{}{+}{k+1}}{\|\gssa{g}{}{+}{k+1}\|}.
\label{eq:normalization}\eeqa Fig.~7 shows the output of a
length-80 equalizer $\ga{g}{k}$ for a 2-source, 4-sensor dynamic
mixing system of order~29, with binary (+1,-1), i.i.d. inputs.
\begin{figure}
\centerline{\psfig{figure=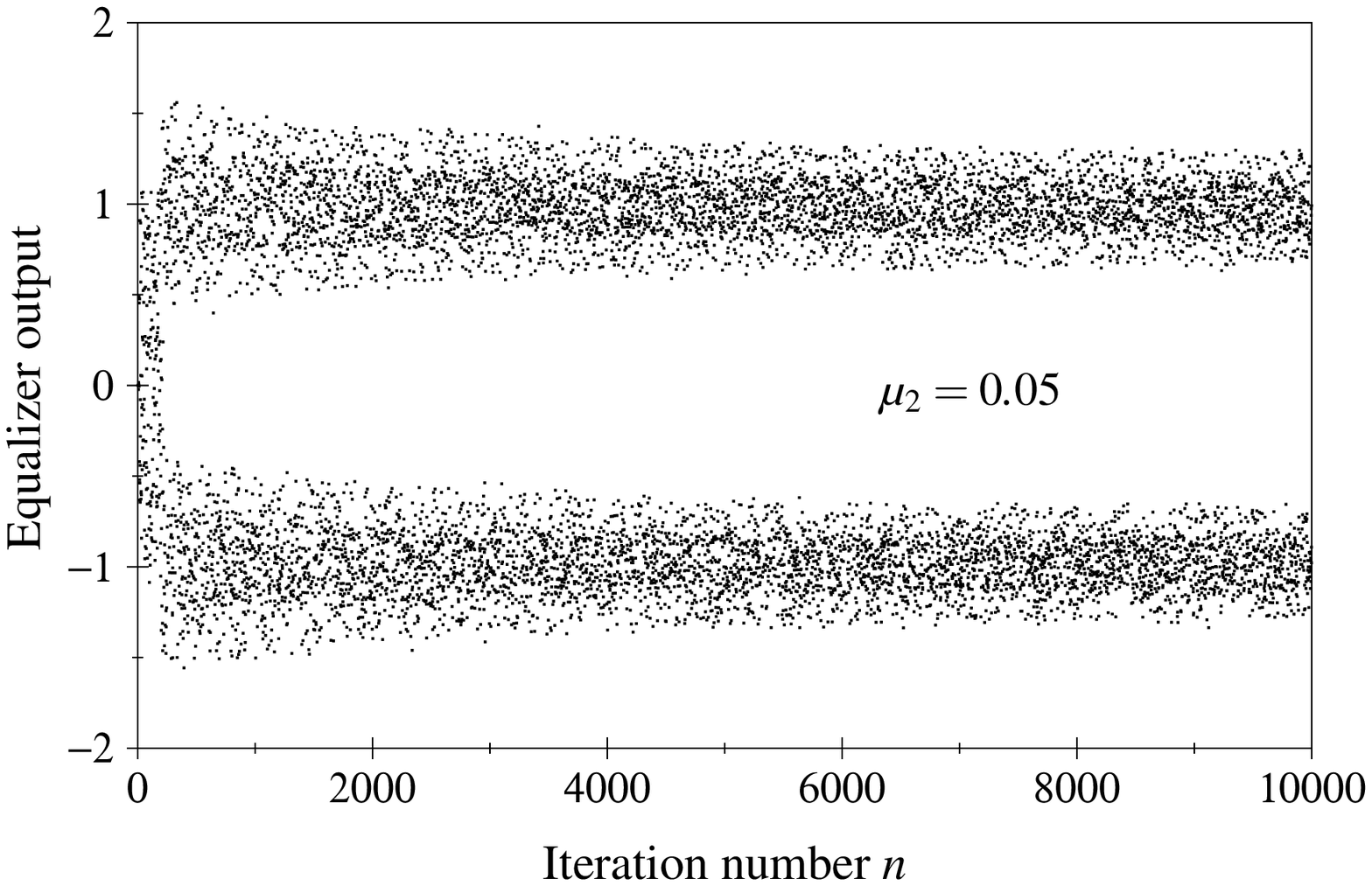,width=16 true cm}}
\caption{Output of the equalizer, adapted with
\pref{eq:Volterra}, \pref{eq:equalizer}, \pref{eq:normalization}.}
\end{figure}

\section{Extensions}
The BSS problem studied in the last example concerns a {\em
dynamic} system with 4 sensors and 2 sources. Its transfer
function can be written as: \beq
\ga{H}{z}=\gss{H}{0}{}+\gss{H}{1}{}z^{-1}+\cdots
+\gss{H}{29}{}z^{-29} \eeq where the $4\times 2$ matrices
$\gss{H}{i}{}$ are the coefficients of its impulse response. The
deconvolution filter is given in the $\cal Z$-domain by the
$4\times 1$ transfer function \beq
\ga{g}{z}=\gss{g}{0}{}+\gss{g}{1}{}z^{-1}+\cdots
+\gss{g}{19}{}z^{-19} \eeq with the $\gss{g}{i}{}$'s being
$4\times 1$ vectors. The separating vector $\gss{g}{}{}$ adapted
with the above algorithm is built from $\ga{g}{z}$ as: \beq
\gss{g}{}{}=\bmx{cccc} \gss{g}{0}{T} & \gss{g}{1}{T} & \cdots
&\gss{g}{19}{T} \emx^{T} \eeq and has dimensions $80\times 1.$
This problem can be re-expressed in the form
$\gss{u}{}{}=\gss{H}{}{}\gss{a}{}{}$ where the matrix
$\gss{H}{}{}$ has the special structure: \beq
\gss{H}{}{}=\bmx{cccccc} \gss{H}{0}{} & \gss{H}{1}{} & \cdots &
\gss{H}{29}{} &\cdots  & \bigcirc \\  & \gss{H}{0}{} & \cdots &
\gss{H}{28}{} & \gss{H}{29}{} & \\ \vdots & \ddots & \ddots &
\ddots & \ddots & \vdots \\ \bigcirc & \cdots & \gss{H}{0}{} &
\cdots & \cdots & \gss{H}{29}{}  \emx \eeq and dimensions
$80\times 98$.

This example shows that the model \pref{eq:u=Ha} can also
describe convolutive mixtures provided an appropriate structure
is imposed on $\gss{H}{}{}.$\footnote{Remark that this demixing
problem is underdetermined, with $M=80<98=N$, contrary to what one
would think by looking at the numbers of sensors and sources.} The
relations between the BSS and blind deconvolution problems are
discussed in \cite{dh00}. That work also considers deriving BSS
algorithms like that of \pref{eq:equivInfomax} that take into
account the rich structure present in $\gss{H}{}{}.$ Similar
adaptive algorithms for blind equalization are derived in
\cite{y98}. In that context, equivariance translates into
robustness to channel ill-conditioning \cite{y98}. The definition
of the natural gradient has been extended to dynamic matrices as
well, giving rise to efficient and fast algorithms \cite{ac98}.
Contrast functions for convolutive mixture separation were
derived in \cite{c96} and later generalized in \cite{mp97}.
Classical multichannel blind deconvolution HOS-based algorithms
\cite{t01} also find application here. One of the few such works
known to cope with the underdetermined case is presented in
\cite{t96}. For more works on convolutive mixture separation,
see, e.g.,
\cite{adcy97,cplm00,cc98,lmv00g,dca97,es97,ln00,lbl97,slj01,t00,t98,t99,zca99}.

Although, as we saw, the noisy case can be studied in the context
of undermodeled systems, a great number of works, dealing
explicitly with noisy mixtures, have also been reported; see,
e.g., \cite{c90,dca98,h99b,mcg97}.

There are applications that involve independent source mixing via
a {\em nonlinear} system. These include satellite channels,
magnetic recording channels, etc \cite{tsj01}. The type of the
nonlinearity suggests the possible separation approaches. In
\cite{yac98} a multilayer perceptron is used as a separator.
Post-nonlinear mixtures, that is, instantaneous linear mixing
systems followed by entrywise zero-memory nonlinearites, are
studied in \cite{tj99a}. In both of these works IT criteria as
those discussed here and related algorithms are employed.
Self-organizing maps are proposed in \cite{phk96}. A more recent
work \cite{tsj01} concerns post-nonlinear convolutive systems.

The case of {\em nonstationary} sources has also been recently
addressed; see, e.g., \cite{cdyc00,cc00,kmo98,ps00}. {\em
Time-varying} mixing systems with stationary sources are studied
in \cite{pn00} where it is shown that for slow variations a
solution is possible by making use of techniques derived for
time-invariant mixtures.

\section{Concluding Remarks}
We gave an overview of the various forms of the BSS problem and
approaches to its solution. Classical and well-established as
well as more recent approaches were discussed in a unifying
framework, putting an emphasis on their connections.

It has to be noted that not all existing BSS works could be
included in such a short course. Only a representative sample was
presented, hopefully sufficiently motivating for further study.
Material left out includes, for example, works on state-space
approaches to BSS (e.g., \cite{zc98}), EM-based algorithms (e.g.,
\cite{bc95}), sparse coding and feature extraction (e.g.,
\cite{hohh98}), independent analysis of subspace components
\cite{car98b,lmv00f}, etc.

The interested readers may consult the relevant (immense)
bibliography. In our list of references, tutorial and review
material is marked in boldface. WWW sites devoted to BSS, with
tutorials, research papers, software and demos, have also been
built and their number is still increasing. A good starting point
is {\tt http://www.media.mit.edu/~paris/ica.html}.

BSS is a topic that can be considered to have reached a certain
maturity, in both its theory and algorithmic schemes. Yet
unanswered questions concerning mainly the understanding of the
behavior and implementation issues of BSS algorithms still exist.
Moreover, new application fields are being discovered and
connections with seemingly unrelated disciplines are brought out,
opening new, promising research directions. One would thus be
justified to feel that BSS will continue to be a research subject
of high interest for the years to come.


\begin{thebibliography}{99}
\bibitem{apr95} U.\ R.\ Abeyratne, A.\ P.\ Petropulu, and J.\ M.\
Reid, ``Higher-order spectra based deconvolution of ultrasound
images," {\em IEEE Trans.\ Ultrasonics, Ferroelectrics, and
Frequency Control,} vol.~42, no.~6, pp.~1064--1095, Nov.~1995.
\bibitem{a98} S.\ Amari, ``Natural gradient works efficiently in
learning," {\em Neural Computation,} vol.~10, pp.~251--276, 1998.
\bibitem{acc97} S.\ Amari, T.-P.\ Chen, and A.\ Cichocki,
``Stability analysis of adaptive blind source separation," {\em
Neural Networks,} vol.~10, no.~8, pp.~1345--1351, 1997.
\bibitem{ac98} S.\ Amari and A.\ Cichocki, {\bf ``Adaptive blind
signal processing -- Neural network approaches,"} pp.~2026--2048
in \cite{procieee98}.
\bibitem{adcy97} S.\ Amari, S.\ C.\ Douglas, A.\ Cichocki, and
H.\ H.\ Yang, ``Multichannel blind deconvolution and equalization
using the natural gradient," {\em Proc.~IEEE Int'l Symp. Wireless
Communications,} pp.~101-107, 1997.
\bibitem{a92} B.\ Arons, ``A review of the cocktail party effect,"
{\em Journal of the American Voice I/O Society,} vol.~12,
pp.~35--50, Jul.~1992.
\bibitem{bmo98} A.\ K.\ Barros, A.\ Mansour, and N.\ Ohnishi,
``Removing artifacts from electrocardiographic signals using
independent components analysis," {\em Neurocomputing,} vol.~22,
no.~1--3, pp.~173--186, Nov.~1998.
\bibitem{b82} R.\ H.\ T.\ Bates, ``Astronomical speckle imaging,"
{\em Physics Reports,} vol.~90, no.~4, pp.~203--297, 1982.
\bibitem{b00} A.\ J.\ Bell,
{\bf ``Information theory, independent-component analysis, and
applications,"} pp.~237--264 in \cite{hay00a}.
\bibitem{bs95} A.\ J.\ Bell and T.\ J.\ Sejnowski, ``An
information-maximization approach to blnd separation and blind
deconvolution," {\em Neural Computation,} vol.~7, pp.~1129--1159,
1995.
\bibitem{bs96} ------, ``Edges are the `independent components' of
natural scenes," {\em Advances in Neural Information Processing
Systems~9 (NIPS-96),} 1996.
\bibitem{b94} S.\ Bellini, ``Bussgang techniques for blind
deconvolution and equalization," pp.~8--59 in \cite{hay94}.
\bibitem{bc95} A.\ Belouchrani and J.-F.\ Cardoso, ``Maximum
likelihood source separation by the expectation-maximization
technique: Deterministic and stochastic implementation," {\em
Proc.~NOLTA-95,} pp.~49--53.
\bibitem{bgr80} A.\ Benveniste, M.\ Goursat, and G.\ Ruget,
``Robust identification of a nonminimum phase system: Blind
adjustment of a linear equalizer in data communications," {\em
IEEE Trans. Automatic Control,} vol.~25, no.~3, pp.~385--399,
June~1980.
\bibitem{b52} J.\ J.\ Bussgang, ``Crosscorrelation functions of
amplitude-distorted Gaussian signals," {\em Technical
Report~216,} MIT Res, Lab. Electronics, pp.~1--14, March~1952.
\bibitem{b97} R.\ Bro, {\bf ``PARAFAC: Tutorial and applications,"} {\em
Chemometrics and Intelligent Laboratory Systems,} vol.~38,
pp.~149--171, 1997.
\bibitem{bsg99} R.\ Bro, N.\ Sidiropoulos, and G.\ B.\ Giannakis,
``A fast least squares algorithm for separating trilinear
mixtures," {\em Proc.~ICA-99,} pp.~289--294.
\bibitem{b95} D.\ S.\ Burdick, {\bf ``An introduction to tensor
products with applications to multiway data analysis,"} {\em
Chemometrics and Intelligent Laboratory Systems,} vol.~28,
pp.~229--237, 1995.
\bibitem{cad96} J.\ A.\ Cadzow, {\bf ``Blind deconvolution via cumulant
extrema,"} {\em IEEE Signal Processing Magazine,} pp.~24--42,
May~1996.
\bibitem{caoliu} X.\ Cao and R.\ Liu, {\bf ``General approach to blind
source separation,"} {\em IEEE Trans.\ Signal Processing,}
vol.~44, no.~3, pp.~562--571, March~1996.
\bibitem{car89} J.-F.\ Cardoso, ``Source separation using higher order
moments," {\em Proc.~ICASSP-89,} pp.~2109--2112.
\bibitem{car90a} ------, ``Localisation et identification par la
quadricovariance," {\em Traitement du Signal,} vol.~7, no.~5,
pp.~397--406, Dec.~1990.
\bibitem{car90b} ------, ``Eigen-structure of the fourth-order cumulant tensor with
application to the blind source separation problem," {\em
Proc.~ICASSP-90,} pp.~2655--2658.
\bibitem{car91} ------, ``Super-symmetric decomposition of the fourth-order cumulant
tensor. Blind identification of more sources than sensors," {\em
Proc.~ICASSP-91,} pp.~3109--3112.
\bibitem{car91b} ------, ``Higher-order narrow-band array
processing," {\em Proc.~Int'l Sig.\ Proc.\ Wkshp on Higher-Order
Statistics,} 1991.
\bibitem{car94} ------, ``On the performance of orthogonal source
separation algorithms," {\em Proc.~EUSIPCO-94.}
\bibitem{car95} ------, ``A
tetradic decomposition of 4th-order tensors: application to the
source separation problem," pp.~375--382 in \cite{mm95}.
\bibitem{car96} ------, {\bf ``Quelques principes de s\'{e}paration de sources,"} in
{\em \'{E}cole des techniques avanc\'{e}es signal-image-parole,}
1996.
\bibitem{car97} ------, ``Infomax and maximum likelihood for blind source
separation," {\em IEEE Signal Processing Letters,} vol.~4, no.~4,
pp.~112--114, April~1997.
\bibitem{car98a} ------, {\bf ``Blind signal separation: statistical
principles,"} {\em Proc.\ IEEE,} vol.~86, no.~10, pp.~2009--2025,
Oct.~1998.
\bibitem{car98b} ------, ``Multidimensional independent component
analysis," {\em Proc.~ICASSP-98,} pp.~1941--1944.
\bibitem{car98c} ------, ``Learning in manifolds: The case of source
separation," {\em Proc.~SSAP-98,} pp.~136--139.
\bibitem{car99} ------, {\bf ``High-order contrasts for independent
component analysis,"} {\em Neural Computation,} vol.~11, no.~1,
pp.~157--192, Jan.~1999.
\bibitem{car00a} ------, {\bf ``Entropic contrasts for source
separation: Geometry and stability,"} pp.~139--189 in
\cite{hay00a}.
\bibitem{car00b} ------, Lecture in {\em GdR Meeting on
Blind Source Separation,} \'{E}cole Nationale Sup\'{e}rieure des
T\'{e}l\'{e}communications (ENST), Paris, 2000.
\bibitem{cc96} J.-F.\ Cardoso and P.\ Comon, {\bf ``Independent component analysis: A survey of
some algebraic methods,"} {\em Proc.~ISCAS-96,} pp.~93--96.
\bibitem{cl96} J.-F.\ Cardoso and B.\ H.\ Laheld, ``Equivariant adaptive source
separation," {\em IEEE Trans.\ Signal Processing,} vol.~44,
no.~12, pp.~3017--3030, Dec.~1996.
\bibitem{cs93} J.-F.\ Cardoso
and A.\ Souloumiac, ``Blind beamforming for non-Gaussian
signals," {\em Proc.~IEE, Part~F,} vol.~140, no.~6, pp.~362--370,
Dec~1993.
\bibitem{cdyc00} C.\ Chang, Z.\ Ding, S.-F.\ Yau, and F.\ H.\ Y.\
Chan, ``Matrix-pencil approach to blind separation of colored
nonstationary signals," {\em IEEE Trans.\ Signal Processing,}
vol.~48, no.~3, pp.~900--907, March~2000.
\bibitem{cplm00} B.\ Chen, A.\ Petropulu, L.\ De Lathauwer, and B.\ De Moor,
``Blind MIMO system identification based on cross-polyspectra,"
{\em Proc.~EUSIPCO-2000.}
\bibitem{cc98} S.\ Choi and A.\ Cichocki, ``Cascade neural
networks for multichannel blind deconvolution," {\em Electronics
Letters,} vol.~34, no.~12, pp.~1186--1187, June~1998.
\bibitem{cc00} ------, ``Blind separation of nonstationary sources
in noisy mixtures," {\em ELectronics Letters,} vol.~36, no.~9,
pp.~848--849, 2000.
\bibitem{ckkv99} A.\ Cichocki, J.\ Karhunen, W.\ Kasprzak, and R.\
Vig\'{a}rio, ``Neural networks for blind separation with unknown
number of sources," {\em Neurocomputing,} vol.~24, pp.~55--93,
1999.
\bibitem{cscos97} A.\ Cichocki, I.\ Sabala, S.\ Choi, B.\ Orsier, and R.\ Szupiluk,
``Self-adaptive independent component analysis for sub-Gaussian
and super-Gaussian mixtures with an unknown number of sources and
additive noise," {\em Proc.~NOLTA-97,} pp.~731--734.
\bibitem{cur94} A.\ Cichocki, R.\ Unbehauen and E.\ Rummert, ``Robust
learning algorithm for blind separation of signals," {\em
Electronics Letters,} vol.~30, no.~17, pp.~1386--1387, Aug.~1994.
\bibitem{c90} P.\ Comon, ``Analyse en composantes ind\'{e}pendantes et identification
aveugle," {\em Traitement du Signal,} vol.~7, no.~3,
pp.~435--450, Dec.~1990.
\bibitem{c93} ------, ``Remarques sur la diagonalisation tensorielle par la methode de
Jacobi,"
{\em Proc.~GRETSI-93,} pp.~125--128.
\bibitem{c94a} ------, ``Independent component analysis, a new concept ?,''
{\em Signal Processing,} vol.~36, pp.~287--314, April~1994.
\bibitem{c94b} ------, {\bf ``Tensor diagonalization, a useful tool
in signal processing,"} {\em Proc.~IFAC-SYSID-94,} pp.~77--82.
\bibitem{c96} ------, ``Contrasts for multichannel blind deconvolution,'' {\em IEEE
Signal Processing Letters,} vol.~3, no.~7, pp.~209--211,
July~1996.
\bibitem{c98} ------, ``Blind channel identification and
extraction of more sources than sensors,'' keynote address in
{\em SPIE Conference,} San Diego, July 19-24 1998, pp.~2--13.
\bibitem{c00a} ------, ``Block methods for channel identification
and source separation," {\em Proc.~ IEEE Symp.\ Adapt.\ Syst.\
Sig.\ Proc. Commun. Control,} pp.~87--92, 2000.
\bibitem{c00b} ------, {\bf ``Tensor decompositions: State of the art and applications,''} keynote address in
{\em IMA Conf. Mathematics in Signal Processing,} Warwick, UK,
Dec.~18-20, 2000.
\bibitem{c01} ------, {\bf ``From source separation to blind
equalization -- Contrast-based approaches,"} {\em
Proc.~ICISP-2001,} pp.~20--32.
\bibitem{cch00} P.\ Comon and P.\ Chevalier, {\bf ``Source separation:
Models, concepts, algorithms and performance,"} pp.~191--236 in
\cite{hay00a}.
\bibitem{clm99} P.\ Comon, L. De Lathauwer, and B. De Moor, ``An algorithm for ICA with 3
sources and 2 sensors,'' {\em Proc.~Sixth Sig. Proc. Workshop on
Higher Order Statistics,} Ceasarea, Israel, June 14--16 1999,
pp.~116--120.
\bibitem{cm96} P.\ Comon and
B.\ Mourrain, ``Decomposition of quantics in sums of powers of
linear forms,'' {\em Signal Processing,} vol.~53, pp.~93--107,
1996.
\bibitem{cb89} R.\ Coppi and S.\ Bolasco (eds.), {\em Multiway
Data Analysis,} Elsevier North-Holland, 1989.
\bibitem{l97} L.\ De Lathauwer, {\em Signal Processing Based on
Multilinear Algebra,} Ph.D.~thesis, K.U.Leuven, Sept.~1997.
\bibitem{lcmv95} L.\ De Lathauwer, P.\ Comon,
B.\ De Moor, and J.\ Vandewalle, `Higher-order power method --
Application in independent component analysis,'' {\em
Proc.~NOLTA-95.}
\bibitem{lmv94} L.\ De Lathauwer, B.\ De Moor, and J.\ Vandewalle,
``Blind source separation by higher-order singular value
decomposition," {\em Proc.~EUSIPCO-94,} pp.~175--178.
\bibitem{lmv00a} ------, ``An algebraic ICA algorithm for 3 sources
and 2 sensors," {\em Proc.~EUSIPCO-2000.}
\bibitem{lmv00b}
------, ``A multilinear singular value decomposition'', {\em SIAM
J.\ Matrix Anal.\ Appl.,} vol.~21, no.~4, pp.~1253--1278,
Apr.~2000.
\bibitem{lmv00c} ------, ``On the best rank-1 and rank-$(R_1, R_2, ..., R_N)$ approximation
and applications of higher-order tensors," {\em SIAM J. Matrix
Anal.\ Appl.,} vol.~21, no.~4, pp.~1324--1342, Apr.~2000.
\bibitem{lmv00d} ------, ``Independent component analysis and (simultaneous) third-order
tensor diagonalization," {\em Internal Report 00-80,} ESAT-SISTA,
K.U.Leuven, 2000 ({\tt
ftp://ftp.esatkuleuven.ac.be/pub/SISTA/delathauwer/reports/ldl-00-80.ps.Z}).
\bibitem{lmv00e} ------, {\bf ``Orthogonal super-symmetric tensor decompositions and independent
component analysis,"} {\em Journal of Chemometrics,} vol.~14,
no.~3, pp.~123--149, May-Jun.~2000.
\bibitem{lmv00f} ------, `Fetal electrocardiogram extraction by blind source
subspace separation,'' {\em IEEE Trans.\ Biomedical Engineering,}
vol.~47, no.~5, pp.~567--572, May~2000.
\bibitem{lmv00g} ------, ``An algebraic approach to blind MIMO identification,''
{\em Proc.~ICA-2000,} pp.~211--214.
\bibitem{dl95} N.\ Delfosse and P.\ Loubaton, ``Adaptive blind
separation of independent sources: A deflation approach," {\em
Signal Processing,} vol.~45, pp.~59--83, 1995.
\bibitem{dhg00} S.\ Dodel, J.\ M.\ Herrmann, and T.\ Geisel,
``Localization of brain activity -- Blind separation of fMRI
data," {\em Neurocomputing,} vol.~32, pp.~71--708, 2000.
\bibitem{d81} D.\ Donoho, ``On minimum entropy deconvolution,"
pp.~565--609 in D.\ F.\ Findley (ed.), {\em Applied Time Series
Analysis II,} Academic Press, 1981.
\bibitem{da00} S.\ C.\ Douglas and S.\ Amari, {\bf ``Natural-gradient
adaptation,"} pp.~13--61 in \cite{hay00a}.
\bibitem{dca97} S.\ C.\ Douglas, A.\ Cichocki, and S.\ Amari,
``Multichannel blind separation and deconvolution of sources with
arbitrary distributions," {\em Proc.~NNSP-97,} pp.~436--445.
\bibitem{dca98} ------, ``Bias removal technique for blind source
separation with noisy measurements," {\em Electronics Letters,}
vol.~34, no.~14, pp.~1379--1380, July~1998.
\bibitem{dh00} S.\ C.\ Douglas and S.\ Haykin, {\bf ``Relationships
between blind deconvolution and blind source separation,"}
pp.~113--145 in \cite{hay00b}.
\bibitem{d97} P.\ Duhamel, {\bf ``Blind multivariable
equalization,"} {\em Proc.~DSP-97,} pp.~13--16.
\bibitem{es97} F.\ Ehlers and H.\ G.\ Schuster, ``Blind separation
of convolutive mixtures and an application in automatic speech
recognition in a noisy environment," {\em IEEE Trans.\ Signal
Processing,} vol.~45, no.~10, pp.~2608--2612, Oct.~1997.
\bibitem{ep98} B.\ Escofier and J.\ Pag\`{e}s, {\em Analyses
Factorielles Simples et Multiples: Objectifs, M\'{e}thodes et
Interpr\'{e}tation,} 3rd. ed., Dunod, Paris, 1998.
\bibitem{fg99} I.\ Fijalkow and P.\ Gaussier, ``Self-organizing
blind mimo deconvolution using lateral-inhibition," {\em
Proc.~ICA-99.}
\bibitem{ghst01a} G.\ B.\ Giannakis, Y.\ Hua, P.\ Stoica, and L.\
Tong (eds.), {\em Signal Processing Advances in Wireless and
Mobile Communications, Vol.~1: Trends in Channel Estimation and
Equalization,} Prentice-Hall, 2001.
\bibitem{ghst01b} ------, {\em Signal Processing Advances in Wireless and
Mobile Communications, Vol.~2: Trends in Single- and Multi-User
Systems,} Prentice-Hall, 2001.
\bibitem{g98} M.\ Girolami, ``An alternative perspective on
adaptive independent component analysis algorithms," {\em Neural
Computation,} vol.~10, pp.~2103--2114, 1998.
\bibitem{g99} ------, {\em Self-Organising
Neural Networks -- Independent Component Analysis and Blind
Source Separation,} Springer-Verlag, 1999.
\bibitem{g80} D.\ N.\ Godard, ``Self-recovering equalization and
carrier tracking in two-dimensional data communication systems,"
{\em IEEE Trans.\ Communications,} vol.~28, no.~11,
pp.~1867--1875, Nov.~1980.
\bibitem{gv96} G.\ H.\ Golub and C.\ F.\ Van Loan, {\em Matrix
Computations,} 3rd ed., Johns Hopkins University Press, 1996.
\bibitem{gc98} O. Grellier and P.\ Comon, ``Blind separation of discrete
sources,'' {\em IEEE Signal Processing Letters,} vol.~5, no.~8,
pp.~212--214, Aug.~1998.
\bibitem{gc99} ------, ``Closed-form
equalization,''{\em Proc.~SPAWC-99,} pp. 219--222.
\bibitem{gc00}  ------, ``Analytical blind discrete source separation,''
{\em Proc.~EUSIPCO-2000.}
\bibitem{gr99} V.\ S.\ Grigorascu and P.\ A.\ Regalia,
``Tensor displacement structures and polyspectral matching''
chap.~9 in {\em Fast Reliable Algorithms for Structured Matrices,}
T.\ Kailath and A.\ H.\ Sayed (eds.), SIAM Publ., Philadelphia,
PA, 1999.
\bibitem{hay94} S.\ Haykin (ed.), {\em Blind Deconvolution,}
Prentice-Hall, 1994.
\bibitem{hay98} S.\ Haykin, {\em Neural Networks: A Comprehensive
Foundation,} 2nd ed., Prentice-Hall, 1999.
\bibitem{hay00a} S.\ Haykin (ed.), {\em Unsupervised Adaptive Filtering,
Vol~1: Blind Source Separation,} John~Wiley\&Sons, 2000.
\bibitem{hay00b} ------, {\em Unsupervised Adaptive Filtering, Vol~2:
Blind Deconvolution,} John~Wiley\&Sons, 2000.
\bibitem{hlk00} T.\ P.\ von Hoff, A.\ G.\ Lindgren, and A.\ N.\
Kaelin, ``Transpose properties in the stability and performance
of the classic adaptive algorithms for blind source separation
and deconvolution," {\em Signal Processing,} vol.~80,
pp.~1807--1822, 2000.
\bibitem{hnb01} R.\ Huez, D.\ Nuzillard, and A.\ Billat,
``Denoising using blind source separation for pyroelectric
sensors," {\em EURASIP J. Appl. Signal Processing,} vol.~2001,
no.~1, pp.~53--65, March~2001.
\bibitem{h98} A.\ Hyv\"{a}rinen, ``New approximations
of differential entropy for independent component analysis and
projection pursuit," {\em Advances in Neural Information
Processing Systems~10 (NIPS-97),} pp.~273--279, MIT~Press, 1998.
\bibitem{h99a} ------, ``Fast and robust fixed-point algorithms for independent component
analysis," {\em IEEE Trans.\ Neural Networks,} vol.~10, no.~3,
pp.~626--634, May~1999.
\bibitem{h99b} ------, ``Gaussian moments for noisy independent component analysis,"
{\em IEEE Signal Processing Letters,} vol.~6, no.~6,
pp.~145--147, June~1999.
\bibitem{h99c} ------, {\bf ``Survey on independent component
analysis,"}
{\em Neural Computing Surveys,} vol.~2, pp.~94--128, 1999.
\bibitem{h99d} ------, ``The fixed-point algorithm and maximum likelihood
estimation for independent component analysis," {\em Neural
Processing Letters,} vol.~10, no.~1, pp.~1--5, Aug.~1999.
\bibitem{hco99} A.\ Hyv\"{a}rinen, R.\ Cristescu, and E.\ Oja, ``A
fast algorithm for estimating overcomplete ICA bases for image
windows," {\em Proc. IJCNN-99.}
\bibitem{ho97} A.\ Hyv\"{a}rinen and E.\ Oja, ``A fast fixed-point algorithm for independent
component analysis," {\em Neural Computation,} vol.~9, no.~7,
pp.~1483--1492, 1997.
\bibitem{ho98} ------, ``Independent component analysis by general nonlinear
Hebbian-like learning rules," {\em Signal Processing,} vol.~64,
pp.~301--313, 1998.
\bibitem{ho00} ------,
{\bf ``Independent component analysis: Algorithms and
applications,"} {\em Neural Networks,} vol.~13, no~4-5,
pp.~411--430, 2000.
\bibitem{hohh98} A.\ Hyv\"{a}rinen, E.\ Oja, P.\ Hoyer, and J.\ Hurri,
"Image feature extraction by sparse coding and independent
component analysis," {\em Proc.~ICPR-98,} pp.~1268--1273.
\bibitem{jh91} C.\ Jutten and J.\ H\'{e}rault, ``Blind separation
of sources, part~I: An adaptive algorithm based on neuromimetic
architecture," {\em Signal Processing,} vol.~24, no.~1,
pp.~1--20, 1991.
\bibitem{j98} C.\ R.\ Johnson {\em et al.,} {\bf ``Blind equalization
using the constant modulus criterion: A review,"} pp.~1927--1950
in \cite{procieee98}.
\bibitem{j00} C.\ R.\ Johnson {\em et al.,} {\bf ``The core of FSE-CMA
behavior theory,"} pp.~13--112 in \cite{hay00b}.
\bibitem{kpo98} J.\ Karhunen, P.\ Pajunen, and E.\ Oja, ``The
nonlinear PCA criterion in blind source separation: Relations
with other approaches," {\em Neurocomputing,} vol.~22, pp.~5--20,
1998.
\bibitem{kmo98} M.\ Kawamoto, K.\ Matsuoka, and N.\ Ohnishi, ``A
method of blind separation for convolved non-stationary signals,"
{\em Neurocomputing,} vol.~22, no.~1--3, pp.~157--171, Nov.~1998.
\bibitem{k99} K.\ Knuth, ``A Bayesian approach to source
separation," {\em Proc.~ICA-99,} pp.~283--288.
\bibitem{kr01a} E.\ Kofidis and P.\ A.\ Regalia, ``Tensor
approximation and signal processing applications," in {\em
Structured Matrices in Operator Theory, Numerical Analysis,
Control, Signal and Image Processing,} Contemporary Mathematics
Series, American Mathematical Society, V.\ Olshevsky (ed.) (to
appear).
\bibitem{kr01b} E.\ Kofidis and P.\ A.\ Regalia, ``On the best
rank-1 approximation of higher-order super-symmetric tensors,"
submitted to {\em SIAM J. Matrix Anal. Appl.}
\bibitem{k77} J.\ B.\ Kruskal, ``Three-way arrays: Rank and
uniqueness of trilinear decompositions, with application to
arithmetic complexity and statistics," {\em Linear Algebra and Its
Applications,} vol.~18, pp.~95--138, 1977.
\bibitem{k89} ------, ``Rank decomposition and uniqueness for
3-way and $N$-way arrays," pp.~8--18 in \cite{cb89}.
\bibitem{kh96} D.\ Kundur and D.\ Hatzinakos, {\bf ``Blind image
deconvolution,"} {\em IEEE Signal Processing Magazine,} May~1996.
See also: {\bf ``Blind image deconvolution revisited,"} {\em IEEE
Signal Processing Magazine,} Nov.~1996.
\bibitem{lac97} J.-L.\ Lacoume, P.-O.\ Amblard, and P.\ Comon,
{\em Statistiques d' Ordre Sup\'{e}rieur Pour le Traitement du
Signal,} Masson, Paris, 1997.
\bibitem{ln00} R.\ H.\ Lambert and C.\ L.\ Nikias, {\bf ``Blind
deconvolution of multipath mixtures,"} pp.~377--436 in
\cite{hay00a}.
\bibitem{l98} T.-W.\ Lee, {\em Independent Component Analysis: Theory and Applications,}
Kluwer Academic Publ., 1998.
\bibitem{lbl97} T.-W.\ Lee, A.\ J.\ Bell, and R.\ H.\ Lambert,
``Blind separation of delayed and convolved sources," {\em
Proc.~NNSP-97.}
\bibitem{lgbs00} T.-W.\ Lee, M.\ Girolami, A.\ J.\ Bell, and T.\ J.\ Sejnowski,
{\bf ``A unifying information-theoretic framework for independent
component analysis,"} {\em Computers and Mathematics with
Applications,} vol.~39, no.~11, pp.~1--21, June~2000.
\bibitem{lgs99} T.-W.\ Lee, M.\ Girolami, and T.\ J.\ Sejnowski,
``Independent component analysis using an extended Infomax
algorithm for mixed subgaussian and supergaussian sources," {\em
Neural Computation,} vol.~11, pp.~417--441, 1999.
\bibitem{lmr01} P.\ Loubaton, \'{E}. Moulines, and P.\ A.\
Regalia, {\bf ``Subspace method for blind identification and
deconvolution,"} pp.~63--112 in \cite{ghst01a}.
\bibitem{mm98} O.\ Macchi and \'{E}. Moreau, ``Source separation
without explicit decorrelation," {\em Proc.~EUSIPCO-98.}
\bibitem{m98} U.\ Madhow, {\bf ``Blind adaptive interference
suppression for Direct-Sequence CDMA,"} pp.~2049--2069 in
\cite{procieee98}.
\bibitem{mr00} M.\ Mboup and P.\ A.\ Regalia, ``A gradient search interpretation
of the super-exponential algorithm,'' {\em IEEE Trans.
Information Theory,} vol.~46, pp.~2731--2734, Nov.~2000.
\bibitem{m87} P.\ McCullagh, {\em Tensor Methods in Statistics,}
Chapman and Hall, 1987.
\bibitem{m91} J.\ M.\ Mendel, {\bf ``Tutorial on higher-order
statistics (spectra) in signal processing and system theory:
Theoretical results and some applications,"} {\em Proc.~IEEE,}
vol.~79, no.~3, pp.~278--305, March~1991.
\bibitem{md99} A.\ Mohammad-Djafari, ``A Bayesian approach to source separation,''
{\em Proc.~MaxEnt-99.}
\bibitem{m00} J.\ R.\ Montalv\~{a}o Filho, {\em \'{E}galisation et
identification de caneaux de communication num\'{e}rique: Une
approche par reconnaissance de formes et m\'{e}lange de
gaussienes,} Ph.D.~Thesis, Universit\'{e} Paris-Sud, Nov.~2000.
\bibitem{mm95} M.\ Moonen and B.\ De Moor (eds.), {\em SVD and
Signal Processing, III: Algorithms, Architectures and
Applications,} Elsevier Science B.V., 1995.
\bibitem{mp97} E.\ Moreau and J.-C.\ Pesquet, ``Generalized
contrasts for multichannel blind deconvolution of linear
systems," {\em IEEE Signal Processing Letters,} vol.~4, no.~6,
pp.~182--183, June~1997.
\bibitem{mcg97} \'{E}. Moulines, J.-F.\ Cardoso, and E.\ Gassiat,
``Maximum likelihood for blind separation and deconvolution of
noisy signals using mixture models," {\em Proc.~ICASSP-97.}
\bibitem{mr93} M.\ K.\ Murray and J.\ W.\ Rice, {\em Differential
Geometry and Statistics,} Chapman and Hall, 1993.
\bibitem{np97} J.-P.\ Nadal and N.\ Parga, ``Redundancy reduction
and independent component analysis: Conditions on cumulants and
adaptive approaches," {\em Neural Computation,} vol.~9,
pp.~1421--1456, 1997.
\bibitem{nm93} C.\ L.\ Nikias and J.\ M.\ Mendel, {\bf ``Signal
processing with higher-order spectra,"} {\em IEEE Signal
Processing Magazine,} pp.~10--37, July~1993.
\bibitem{od98} D.\ Obradovic and G.\ Deco, ``Information
maximization and independent component analysis: Is there a
difference?," {\em Neural Computation,} vol.~10, pp.~2085--2101,
1998.
\bibitem{oit98} S.\ Ohno, Y.\ Inouye, and K.\ Tanebe, ``A
numerical algorithm for implementing the single-stage
maximization criterion for multichannel blind deconvolution,"
{\em Proc.~EUSIPCO-98.}
\bibitem{od00} G.\ C.\ Orsak and S.\ C.\ Douglas, ``Code-length-based universal extraction
for blind signal separation," {\em Proc.~ICASSP-2000,}
pp.~3422--3425.
\bibitem{phk96} P.\ Pajunen, A.\ Hyv\"{a}rinen and J.\ Karhunen,
``Non-linear blind source separation by self-organizing maps,"
{\em Proc.~Int. Conf. on Neural Information Systems,}
pp.~1207--1210, 1996.
\bibitem{pk99} P.\ Pajunen and J.\ Karhunen, ``Least-squares
methods for blind source separation based on nonlinear PCA," {\em
Int'l J. Systems Science,} April~1999.
\bibitem{p00} C.\ B.\ Papadias, {\bf ``Blind separation of independent
sources based on multiuser kurtosis optimization criteria,"}
pp.~147--179 in \cite{hay00b}.
\bibitem{p84} A.\ Papoulis, {\em Probability, Random Variables
and Stochastic Processes,} McGraw-Hill, 1984.
\bibitem{pn00} N.\ Parga and J.-P.\ Nadal, ``Blind source
separation with time-dependent mixtures," {\em Signal
Processing,} vol.~80, pp.~2187--2194, 2000.
\bibitem{ps00} L.\ Parra and C.\ Spence, ``Convolutive blind
separation of non-stationary sources," {\em IEEE Trans.\ Speech
and Audio Processing,} vol.~8, no.~3, pp.~320--327, 2000.
\bibitem{p01} D.-T.\ Pham, ``Blind separation of instantaneous
mixture of sources via the Gaussian mutual information
criterion," {\em Signal Processing,} vol.~81, pp.~855--870, 2001.
\bibitem{pxf00} J.\ C.\ Principe, D.\ Xu, and J.\ W.\ Fisher III,
{\bf ``Information-theoretic learning,"} pp.~265--319 in
\cite{hay00a}.
\bibitem{r99} P.\ A.\ Regalia, ``On the equivalence between the
Godard and Shalvi-Weinstein schemes of blind equalization,'' {\em
Signal Processing,} vol.~73, pp.~185--190, 1999.
\bibitem{rk00a} P.\ A.\ Regalia and E.\ Kofidis, ``The
higher-order power method revisited: Convergence proofs and
effective initialization," {\em Proc.~ICASSP-2000.}
\bibitem{rk00b} ------, ``A ``unimodal" blind equalization
criterion," {\em Proc.~EUSIPCO-2000.}
\bibitem{rk01} ------, ``Global
convergence of Hyv\"{a}rinen's fixed-point algorithms," submitted
to {\em IEEE Trans. Neural Networks.}
\bibitem{rm99a} P.\ A.\ Regalia and M.\ Mboup,
``Undermodeled equalization: A characterization of stationary
points for a family of blind criteria,'' {\em IEEE Trans.\ Signal
Processing,} vol.~47, no.~3, pp.~760--770, March~1999.
\bibitem{rm99b} ------, ``Equalization in noisy multi-user
channels," {\em Proc.~IEEE/EURASIP Int'l Wkshop.\ Nonlinear Signal
and Image Processing,} 1999.
\bibitem{sj98} P.\ Schniter and C.\ R.\ Johnson, Jr.,
``Minimum-entropy blind acquisition/equalization for uplink
DS-CDMA," {\em Proc.~36th Allerton Conf. Commun., Control, and
Computing,} 1998.
\bibitem{sw90} O.\ Shalvi and E.\ Weinstein, ``New criteria for
blind deconvolution of nonminimum phase systems (channels)," {\em
IEEE Trans.\ Information Theory,~} vol.~36, no.~2, pp.~312--321,
March~1990.
\bibitem{sw93} ------, ``Super-exponential methods for blind
deconvolution," {\em IEEE Trans.\ Information Theory,} vol.~39,
no.~2, pp.~504--519, March~1993.
\bibitem{sw94} ------, {\bf ``Universal methods for blind
deconvolution,"} pp.~121--180 in \cite{hay94}.
\bibitem{sg93} S.\ Shamsunder and G.\ B.\ Giannakis, ``Modeling of
non-Gaussian array data using cumulants: DOA estimation of more
sources with less sensors," {\em Signal Processing,} vol.~ 30,
pp.~ 279--297, 1993.
\bibitem{s98} J.\ Sheinvald, ``On blind beamforming for multiple
non-Gaussian signals and the constant-modulus algorithm," {\em
IEEE Trans.\ Signal Processing,} vol.~46, no.~7, pp.~1878--1885,
July~1998.
\bibitem{sb00} N.\ Sidiropoulos and R.\ Bro, ``On communication
diversity for blind identifiability and the uniqueness of
low-rank decomposition of $N$-way arrays," {\em
Proc.~ICASSP-2000.}
\bibitem{sb01} ------, {\bf ``PARAFAC techniques
for signal separation,"} pp.~131--179 in \cite{ghst01b}.
\bibitem{slj01} C.\ Simon, P.\ Loubaton, and C.\ Jutten,
``Separation of a class of convolutive mixtures: A contrast
function approach," {\em Signal Processing,} vol.~81,
pp.~883--887, 2001.
\bibitem{s97} P.\ S.\ Smaragdis, {\bf {\em Information Theoretic
Approaches to Source Separation,}} MSc Thesis, MIT Media Lab.,
1997.
\bibitem{procieee98} Special issue on {\bf ``Blind system identification and
estimation,"} {\em Proc.~IEEE,} vol.~86, no.~10, Oct.~1998.
\bibitem{s00} M.\ Stetter {\em et al.,} ``Principal component
analysis and blind separation of sources for optical imaging of
intrinsic signals," {\em NeuroImage,} vol.~11, no.~5~I,
pp.~482--490, 2000.
\bibitem{sm98} B.\ Stoll and \'{E}. Moreau, ``Nonlinear
constrained optimization using Lagrangian approach for blind
source separation," {\em Proc.~EUSIPCO-98.}
\bibitem{tj99a} A.\ Taleb and C.\ Jutten, ``On underdetermined
source separation," {\em Proc.~ICASSP-99.}
\bibitem{tj99b} ------, ``Source separation in
post-nonlinear mixtures," {\em IEEE Trans.\ Signal Processing,}
vol.~47, no.~10, pp.~2807--2820, Oct.~1999.
\bibitem{tsj01} A.\ Taleb, J.\ Sol\'{e}, and C.\ Jutten,
``Quasi-nonparametric blind inversion of Wiener systems," {\em
IEEE Trans.\ Signal Processing,} vol.~49, no.~5, pp.~917--924,
May~2001.
\bibitem{t96} L.\ Tong, ``Identification of multichannel MA
parameters using higher-order statistics," {\em Signal
Processing,} vol.~53, pp.~195--209, 1996.
\bibitem{til93} L.\ Tong, Y.\ Inouye, and R.\ Liu,
``Waveform-preserving blind estimation of multiple independent
sources," {\em IEEE Trans.\ Signal Processing,} vol.~41, no.~7,
pp.~2461--2470, July~1993.
\bibitem{tlsh91} L.\ Tong, R.\ Liu, V.\ C.\ Soon, and Y.-F.\
Huang, ``Indeterminacy and identifiability of blind
identification," {\em IEEE Trans.\ Circuits and Systems,}
vol.~38, no.~5, pp.~499--509, May~1991.
\bibitem{t00} K.\ Torkkola, {\bf ``Blind separation of delayed and
convolved sources,"} pp.~321--375 in \cite{hay00a}.
\bibitem{ta83} J.\ R.\ Treichler and B.\ G.\ Agee, ``A new
approach to multipath correction of constant modulus signals,"
{\em IEEE Trans.\ Acoust., Speech, Signal Processing,} vol.~31,
no.~2, pp.~459--472, Feb.~1983.
\bibitem{tlh98} J.\ R.\ Treichler, M.\ G.\ Larimore, and J.\ C.\
Harp, {\bf ``Practical blind demodulators for high-order QAM
signals,"} pp.~1907--1926 in \cite{procieee98}.
\bibitem{t98} J.\ K.\ Tugnait, ``On blind separation of
convolutive mixtures of independent linear signals in unknown
additive noise," {\em IEEE Trans.\ Signal Processing,} vol.~46,
no.~11, pp.~3117--3123, Nov.~1998.
\bibitem{t99} ------, ``Adaptive blind separation of convolutive
mixtures of independent linear signals," {\em Signal Processing,}
vol.~ 75, pp.~139--152, 1999.
\bibitem{t01} ------, {\bf ``Channel estimation and
equalization using higher-order statistics,"} pp.~1--39 in
\cite{ghst01a}.
\bibitem{v98} A.-J.\ van der Veen, {\bf ``Algebraic methods for
deterministic blind beamforming,"} pp.~1987--2008 in
\cite{procieee98}.
\bibitem{v01} ------, {\bf ``Algebraic constant modulus
algorithms,"}
pp.~89--130 in \cite{ghst01b}.
\bibitem{vp96} A.-J.\ van der Veen and A.\ Paulraj, ``An
analytical constant modulus algorithm," {\em IEEE Trans.\ Signal
Processing,} vol.~44, no.~5, pp.~1136--1155, May~1996.
\bibitem{w90} H.-S.\ Wu, ``Minimum entropy deconvolution for
restoration of blurred two-tome images," {\em Electronics
Letters,} vol.~26, no.~15, pp.~1183--1184, July~1990.
\bibitem{yi00} T.\ Yamaguchi and K.\ Itoh, ``An algebraic solution
to independent component analysis," {\em Optics Communications,}
vol.~178, pp.~59--64, May~2000.
\bibitem{y98} H.\ H.\ Yang, ``On-line blind equalization via
on-line blind separation," {\em Signal Processing,} vol.~68,
no.~3, pp.~271--281, 1998.
\bibitem{ya97} H.\ H.\ Yang and S.\ Amari, ``Adaptive on-line
learning algorithms for blind separation: Maximum entropy and
minimum mutual information," {\em Neural Computation,} vol.~9,
pp.~1`457--1482, 1997.
\bibitem{yac98} H.\ H.\ Yang, S.\ Amari, and A.\ Cichocki,
``Information-theoretic approach to blind separation of sources in
non-linear mixture," {\em Signal Processing,} vol.~64,
pp.~291--300, 1998.
\bibitem{zac99} L.-Q.\ Zhang, S.\ Amari, and A.\ Cichocki,
``Natural gradient approach to blind separation of over- and
under-complete mixtures," {\em Proc.~ICA-99.}
\bibitem{zc98} L.-Q.\ Zhang and A.\ Cichocki,
``Blind deconvolution/equalization using state-space models,"
{\em Proc.~NNSP'98,} pp.~123--131.
\bibitem{zca99} L.\ Q.\ Zhang, A.\ Cichocki, and S.\ Amari,
Multichannel blind deconvolution of non-minimum phase systems
using information backpropagation," {\em Proc.~ICONIP-99,}
pp.~210--216.
\end{thebibliography}
\end{document}